\documentclass{article}

 \usepackage[main, final]{neurips_2026}

\usepackage[utf8]{inputenc} 
\usepackage[T1]{fontenc}    
\usepackage{hyperref}       
\usepackage{url}            
\usepackage{booktabs}       
\usepackage{amsfonts}       
\usepackage{nicefrac}       
\usepackage{microtype}      
\usepackage{xcolor}         

\usepackage{graphicx}
\usepackage{amsmath}
\usepackage{wrapfig}
\usepackage{tikz}
\usepackage{caption}
\usepackage{algorithm}
\usepackage{algpseudocode}
\usepackage{subcaption}
\usepackage{lineno}

\usepackage{cleveref}
\crefname{section}{Sec.}{Secs.}
\Crefname{section}{Sec.}{Secs.}
\crefname{figure}{Fig.}{Figs.}
\Crefname{figure}{Fig.}{Figs.}
\crefname{table}{Tab.}{Tabs.}
\Crefname{table}{Tab.}{Tabs.}
\crefname{equation}{Eq.}{Eqs.}
\Crefname{equation}{Eq.}{Eqs.}

\newcommand{\etal}{\textit{et al.}}

\title{MonoPhysics: Estimating Geometry, Appearance, and Physical Parameters from Monocular Videos}

\author{%
  Daniel Rho$^1$\hspace{0.9em}Jun Myeong Choi$^1$\hspace{0.9em}Matthew Thornton$^1$\hspace{0.9em}Biswadip Dey$^2$\hspace{0.9em}Roni Sengupta$^1$\\
  \\
  $^1$University of North Carolina at Chapel Hill \quad $^2$Meta
}

\begin{document}

\maketitle

\begin{center}
    \vspace{-30pt}
    \nolinenumbers
    \includegraphics[width=0.95\textwidth]{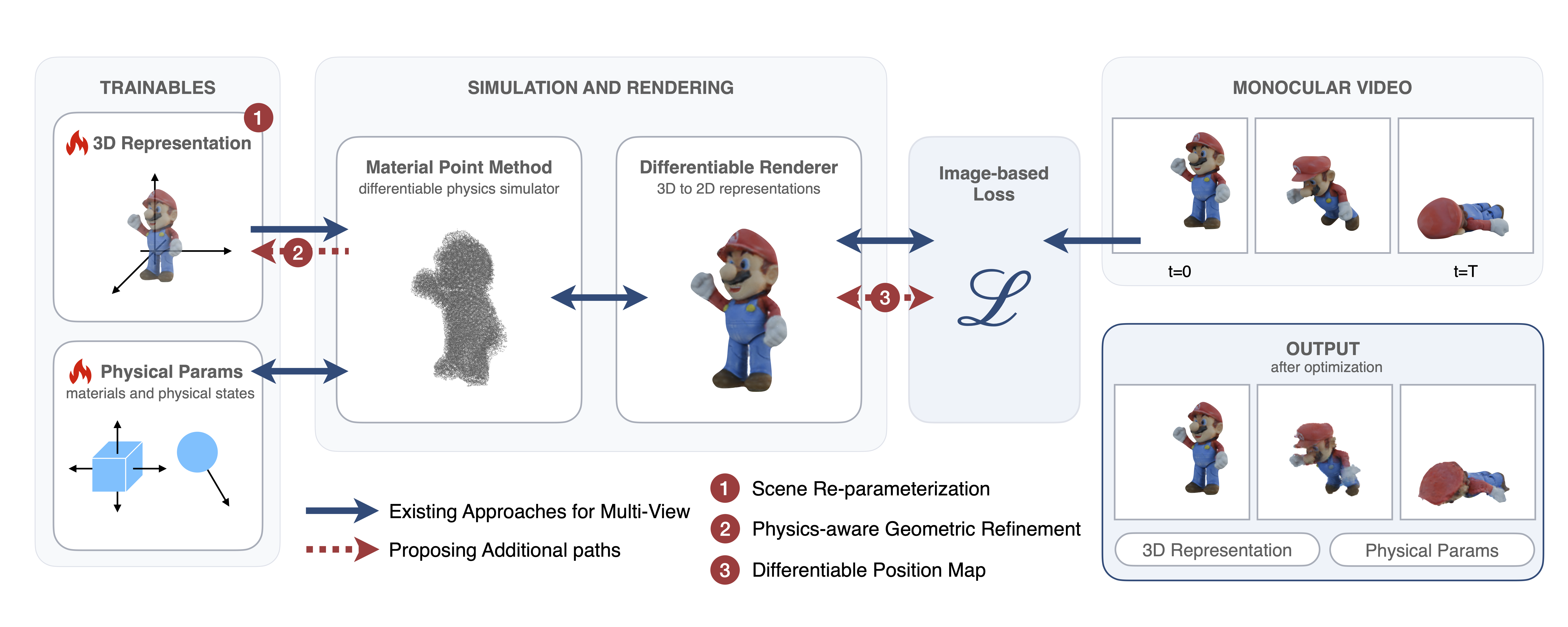}
    \vspace{-10pt}
    \captionof{figure}{
        Overview of \textit{MonoPhysics}.
        A trainable 3D representation and physical parameters are simulated and rendered, and supervised by image-based losses $\mathcal{L}$ against a monocular video.
        Solid blue arrows indicate optimization paths shared with existing multi-view methods, while red dashed arrows indicate the additional paths proposed in this work.
        \textit{MonoPhysics} introduces three contributions, numbered in the figure: scene re-parameterization, physics-aware geometric refinement, and differentiable position map.
    }
    \label{fig:teaser}
    \linenumbers
\end{center}

\begin{abstract}
  Existing inverse physics methods recover physical parameters from multi-view videos, where geometric constraints across views resolve scale and 3D structure.
  In monocular settings, however, such constraints are absent, leading to severe scale ambiguity, inaccurate geometry, and weak coupling between appearance optimization and physical simulation.
  To address these challenges, we propose MonoPhysics, a framework for monocular inverse physics estimation of deformable objects that jointly optimizes geometry, appearance, and physical parameters using a differentiable simulator and 3D Gaussian Splatting.
  Our key contribution is removing the multi-view capture requirement of existing methods, a necessary step toward handling in-the-wild video.
  MonoPhysics introduces three visual-physical bridges: scene re-parameterization, physics-aware geometry refinement, and a differentiable position map.
  We evaluate on Vid2Sim, real-world captures, and a new dataset of elastic and plasticine objects that we introduce.
  MonoPhysics outperforms monocular baselines in future prediction and recovers Young's modulus on Vid2Sim with accuracy comparable to a multi-view baseline.
  Code and data are available at \url{https://daniel03c1.github.io/MonoPhysics/}.
\end{abstract}

\section{Introduction}
\label{sec:introduction}

Recovering the physical properties of objects from video, such as material stiffness and deformation behavior, is important for robotic manipulation of deformable objects, as well as for building digital twins and virtual environments~\cite{jiang2025phystwin}.
Recent advances in differentiable physics simulation~\cite{Hu2020DiffTaichi:,murthy2021gradsim} and neural 3D representations~\cite{NeRF,kerbl3Dgaussians} have enabled \emph{inverse physics}: recovering physical parameters by optimizing through differentiable simulations to match observed motion.
Existing inverse physics methods~\cite{li2023pacnerf,SpringGaus,GIC,MASIV,vid2sim} typically rely on synchronized multi-view captures, where geometric constraints across views resolve 3D structure and absolute scale.
However, multi-view rigs are unavailable for the most common sources of dynamic video, including most in-the-wild captures.
Removing the multi-view capture requirement is therefore an important step toward physical system identification from such videos.

Multi-view methods~\cite{li2023pacnerf,SpringGaus,GIC,MASIV,vid2sim} rely on triangulation to resolve 3D structure and absolute scale before estimating physical parameters.
The physics optimization therefore operates on geometry already constrained by visual evidence.

The monocular setting is significantly more challenging, as geometry must be inferred without multi-view constraints and is inherently ambiguous in scale.
While recent 3D foundation models~\cite{sam3dteam2025sam3d3dfyimages} can recover plausible geometry from a single image and provide a strong starting point, their outputs remain unreliable in occluded regions and lack absolute scale, both critical for physical simulation.
Prior inverse physics methods typically decouple geometry reconstruction and physical parameter estimation, assuming a reliable 3D representation can be obtained from multi-view inputs~\cite{li2023pacnerf,SpringGaus,GIC,MASIV}.
In the monocular setting, this decoupling breaks down, often requiring joint optimization of geometry, appearance, and physical parameters.

We present \textit{MonoPhysics}, a framework for monocular inverse physics of deformable objects based on differentiable MPM simulation and 3D Gaussian Splatting.
It is initialized from a 3D foundation model~\cite{sam3dteam2025sam3d3dfyimages} and jointly refined under physical and visual supervision.
Our key contribution is removing the multi-view capture requirement for accurate inverse physics, a step toward in-the-wild use.
We use 3D Gaussians for both rendering and simulation: each Gaussian serves as a rendering primitive and as a material point for simulation.
Thus, the visual and physical sides of the pipeline are coupled through differentiable rendering and simulation.
In the monocular setting, we identify several connections between them that remain weak or missing, preventing visual cues from fully guiding the underlying physical parameters.

In this work, we propose three components that bridge the visual and physical sides of the pipeline.
First, scene re-parameterization separates global scale from geometry.
Since per-particle position updates produce only local corrections, a single learnable scalar helps aggregate a coherent global scale signal across all particles.
Second, physics-aware geometry refinement dynamically computes and optimizes per-particle volumes, and uses both visual and physical importance to guide Gaussian relocation during optimization.
Third, a differentiable position map provides gradients on the rendered positions of Gaussians, enabling direct position-based supervision that standard 3DGS rendering cannot provide.
We optimize geometry, appearance, and physical parameters using a combination of rendering, optical flow, and silhouette losses, together with a particle distribution regularizer.

We evaluate MonoPhysics on the Vid2Sim benchmark~\cite{vid2sim}, our new synthetic dataset of elastic and plasticine objects, and real captures from SpringGaus~\cite{SpringGaus}, comparing against existing methods~\cite{vid2sim,GIC,SpringGaus,li2023pacnerf}.
Our approach achieves the best future prediction and the lowest Young's modulus error among monocular methods on all three datasets.
Notably, using only a single camera, it reaches Young's modulus accuracy on Vid2Sim comparable to that of a multi-view method.
\section{Related Works}
\label{sec:related_works}

\subsection{Inverse Physics from Video}
\label{ssec:inverse_physics}
Estimating physics from video sequences has been studied extensively~\cite{li2023pacnerf,UniPhy}.
Differentiable simulation frameworks~\cite{Hu2020DiffTaichi:,murthy2021gradsim} enable gradient-based optimization of physical parameters by differentiating through the simulation.
This allows observed object behavior to directly supervise material properties, initial velocities, and other physical parameters.
PAC-NeRF~\cite{li2023pacnerf} combines differentiable MPM simulation with NeRF to optimize geometry and physical parameters from multi-view video.
SpringGaus~\cite{SpringGaus} integrates a spring-mass model into 3D Gaussian Splatting for reconstruction and simulation of elastic objects.
GIC~\cite{GIC} applies point-wise losses on particles from a learned 4D representation to estimate physical parameters.
Vid2Sim~\cite{vid2sim} uses feed-forward prediction for initialization, followed by lightweight optimization with a mesh-free differentiable simulator~\cite{simplicits}.
MASIV~\cite{MASIV} introduces learnable neural constitutive models~\cite{NCLaws} that remove the need for hand-crafted constitutive laws, enabling material-agnostic system identification.
These methods benefit from multi-view constraints or direct 4D guidance, which reduces the problem to estimating physical parameters from already resolved geometry.
In the monocular setting, however, these constraints are absent.

\subsection{Monocular and Sparse-View Inverse Physics Estimation}
\label{ssec:monocular}
Monocular video poses unique challenges for inverse physics.
Without multi-view constraints, scale is ambiguous, geometry is inaccurate, and appearance provides only limited supervision.
Prior monocular or limited-view methods partially address these issues.
ProJo4D~\cite{rho2025projo4d} improves parameter estimation and prediction in sparse-view settings, but still relies on multi-view input.
NeuPhysics~\cite{qiao2022neuphysics} learns a 4D representation from monocular video and estimates physical parameters from the reconstructed trajectories.
However, it separates representation learning from parameter estimation and thus assumes accurate 4D reconstruction.
PPR~\cite{PPR} resolves scale for articulated bodies, but relies on priors that do not apply to general deformable objects.
Gao~\etal~\cite{gao2025seeing} estimate the physical parameters of thin deformable objects from monocular video, assuming full surface visibility in the first frame.
FluidNexus~\cite{gao2025fluidnexus} reconstructs and predicts fluid dynamics from a single video, but is limited to fluids.
In contrast, our framework directly addresses monocular-specific challenges, such as ambiguous scale and inaccurate geometry, using physics-derived signals rather than object-specific priors or multi-view constraints.

\subsection{Physics-Informed Scene Understanding}
\label{ssec:physics_informed}
Physical constraints and priors have been used to improve static scene reconstruction and infer object structure.
PhyRecon~\cite{ni2024phyrecon} integrates differentiable rendering with particle-based physics simulation to produce physically stable static scene reconstructions.
Guo~\etal~\cite{guo2024physically} optimize for physical compatibility to obtain plausible static objects from a single image.
PhySIC~\cite{ym2025physic} uses physical contact constraints for human-scene alignment from a single image.
BrickGPT~\cite{pun2025brickgpt} uses physical stability constraints to produce buildable brick structures.
TopoGaussian~\cite{xiong2025topogaussian} and Structure from Collision~\cite{SfC} estimate interior structure from multi-view input.
However, these methods either operate on static scenes or require multi-view input.
Closer to our setting, Physics-Informed Deformable Gaussian Splatting~\cite{hong2025physics} uses physics as regularization for monocular 4D reconstruction and learns time-varying material parameters, but does not perform full inverse estimation with forward prediction.
In contrast, our work uses physics as a corrective signal specifically for monocular inverse estimation of dynamic deformable objects.

\section{Method}
\label{sec:method}
We first formulate the problem overview (\cref{ssec:overview}).
Our three contributions are bridges between the visual and physical sides of the pipeline (\cref{sec:introduction}): a global scale scalar (\cref{ssec:scale}), physics-aware geometry refinement (\cref{ssec:mcmc}), and a differentiable position map (\cref{ssec:posmap}).
Loss functions and the full optimization procedure follow in \cref{ssec:optimization}.

\subsection{Problem Formulation and Overview}
\label{ssec:overview}
The input is a monocular RGB video of a deformable object interacting with its environment, e.g., an object falling onto a table.
We assume known camera parameters (intrinsics and extrinsics), a known ground plane, and a known constitutive model (e.g., Neo-Hookean elasticity), consistent with prior inverse physics methods~\cite{li2023pacnerf,SpringGaus,GIC}.
We also require a foreground alpha mask for each frame.
When ground-truth masks are unavailable, an off-the-shelf segmentation method can be used.
Our goal is to jointly recover the object's geometry, appearance, and material parameters from these inputs.
For 3D representation, we use 3D Gaussian Splatting (3DGS)~\cite{kerbl3Dgaussians}, which integrates with particle-based simulations such as the Material Point Method (MPM).
Each Gaussian has position $\mathbf{x}_i$, color $\mathbf{c}_i$, opacity $o_i$, and covariance $\boldsymbol{\Sigma}_i$ for appearance, and physical volume $V_i$.
The initial physical state is represented by the initial velocity $\mathbf{v}_0$. 

We use an off-the-shelf monocular 3D reconstruction method~\cite{sam3dteam2025sam3d3dfyimages} to produce an initial set of 3D Gaussians from the first frame, then sample MPM particles from them.
We use a differentiable MPM simulator~\cite{Hu2020DiffTaichi:} to advance them through time under the current physical parameters, producing deformed representations at each frame.
At each frame, the color of every pixel $C$ is rendered via 3D Gaussian Splatting using alpha compositing~\cite{kerbl3Dgaussians}:
\begin{equation}
\label{eq:3dgs_render}
    C = c_{bg}\prod_{i}\bigl(1 - \alpha_i\bigr) + \sum_{i} \alpha_i\,\mathbf{c}_i \prod_{j < i}\bigl(1 - \alpha_j\bigr),
\end{equation}
where $\mathbf{c}_i$ is the color of Gaussian $i$, $\alpha_i$ is the product of the learned opacity and the Gaussian response of particle $i$ at the pixel, and $c_{bg}$ is the background color of that pixel.
Losses are backpropagated through the differentiable simulation to update all parameters jointly, as detailed in \cref{ssec:optimization}.

\subsection{Scale Estimation through Reparameterization}
\label{ssec:scale}

\setlength\intextsep{-2pt}
\begin{wrapfigure}{r}{0.47\textwidth}
    \centering
    \includegraphics[width=\linewidth,trim=6 8 6 6,clip]{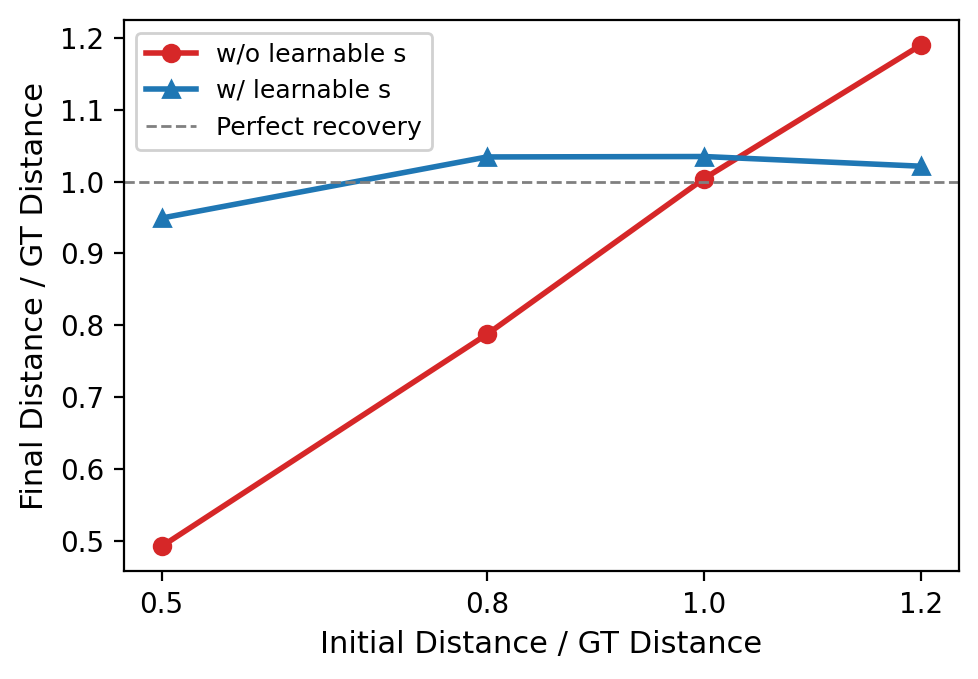}
    \caption{
        Effect of the learnable scale factor $s$ on the recovered center-of-mass distance across different initial scales.
        Without $s$, the optimization cannot correct errors in the initial scale estimate, whereas with $s$, our method recovers a near-correct scale across tested initializations.
    }
    \label{fig:abl_scale}
\end{wrapfigure}

Monocular 3D reconstruction recovers scene geometry only up to an unknown scale factor, as points along the same projection ray are indistinguishable from the camera's perspective.
For physics, however, absolute scale governs world-space positions, velocities, contact timing, volumes, and stress magnitudes, so all downstream physical quantities depend on it.
Scale error therefore propagates into every estimated parameter.
\setlength\intextsep{12pt plus 2pt minus 2pt} 

Physical dynamics can resolve this ambiguity.
However, naively optimizing per-particle positions cannot recover the global scale.
Each particle's gradient mixes local position corrections with the global scale signal, so independent per-particle updates produce local adjustments rather than a coherent change in the object's distance from the camera (\cref{fig:abl_scale}).

Existing methods store particles directly in world space~\cite{li2023pacnerf,SpringGaus,GIC,MASIV}.
We instead maintain particles in camera space and introduce a single learnable scalar $s$ that controls the global scale.
Scaling in camera space moves particles along their projection rays, leaving 2D projections unchanged but changing all physical quantities.
Since physics simulation requires world-space coordinates, we transform using the known camera extrinsics:
\begin{align}
\label{eq:c2w}
    \mathbf{x}_i^w &= \mathbf{R}\,(s \cdot \mathbf{x}_i) + \mathbf{t},
\end{align}
where $\mathbf{R}, \mathbf{t}$ are the camera-to-world rotation and translation, $\mathbf{x}_i$ is the camera-space position. 
Because every particle's physics depends on $s$, image-space losses backpropagated through the differentiable simulation aggregate a coherent global signal for scale from all particles and all frames.
This gives $s$ a clean optimization path that per-particle degrees of freedom cannot provide (\cref{fig:abl_scale}).

\subsection{Physics-Aware Geometry Refinement}
\label{ssec:mcmc}
Accurately estimating geometry from a single camera is difficult: unobserved regions remain inherently ambiguous even with 3D geometric priors from foundation models.
Our key idea is that a differentiable physics simulation (MPM) provides complementary signals to refine geometry, since physical behavior constrains regions that visual observation alone cannot resolve.
In neural field-based representations~\cite{li2023pacnerf,SfC,Kaneko_2024_CVPR}, geometry can be refined by optimizing continuous density or occupancy fields.
Particle-based representations lack this continuous structure, so geometry is instead captured through per-particle volumes.

\noindent \textbf{Per-particle physical volume.}
Existing works that incorporate physics into 3DGS representations~\cite{GIC,MASIV} assign fixed per-particle volumes, computed once at initialization.
In standard MPM initialization, these volumes are chosen to satisfy a local partition-of-unity over the simulation grid: the total volume contributed by particles to each grid node matches the cell volume $\Delta x^3$.
More specifically, the total volume contributed to each grid node equals the cell volume $\Delta x^3$:
\begin{equation}
\label{eq:partition_of_unity}
\sum_i V_i w_{i\ell} = \Delta x^3,
\end{equation}
where $\Delta x$ denotes cell edge size, and $w_{i\ell}$ is the contribution of particle $i$ to grid node $\ell$.
During initialization, these per-particle physical volumes are determined.
However, keeping volumes fixed throughout optimization prevents the representation from adapting its geometry and total volume.

We instead make volumes adaptive by maintaining the partition-of-unity (\cref{eq:partition_of_unity}), inducing volumes from the current particle distribution at each step.
We compute them by iteratively refining per-particle volume to meet the partition of unity for $T$ iterations.
Further details are provided in the supplementary.

\noindent \textbf{Physics-informed particle management.}
Adaptive volumes alone cannot add particles where geometry is under-resolved, nor remove particles wasted on negligible regions.
We therefore complement adaptive volumes with a particle management scheme that relocates particles during optimization.
Prior visual-only methods select particles using visual importance only~\cite{kerbl3Dgaussians,Lee_2024_CVPR,3DGS_MCMC}.
We propose using physical importance in addition to visual importance.
We define a visual importance $p_i^{vis}$ from the rendering footprint, a physical importance $p_i^{phys}$ from the physical volume, and the overall importance $p_i$ as the average of the two:
\begin{equation}
\label{eq:importance}
    p_i \propto \tfrac{1}{2}p_i^{vis} + \tfrac{1}{2}p_i^{phys}, \quad p_i^{vis} \propto \det(\boldsymbol{\Sigma}_i)^{1/2} \cdot o_i, \quad p_i^{phys} \propto V_i,
\end{equation}
where $\boldsymbol{\Sigma}_i$ is the 3D covariance matrix and $o_i$ is the learned opacity of particle $i$.
3DGS-MCMC~\cite{3DGS_MCMC} relocates low-opacity Gaussians to high-opacity ones and adjusts their opacity and scale so that the rendered image is approximately preserved.
We reuse this relocation step but select which particles to remove and split using the importance in \cref{eq:importance} instead of opacity alone.
Specifically, we periodically remove particles with the lowest visual importance $p_i^{\text{vis}}$ or physical importance $p_i^{\text{phys}}$, and replace them by splitting those with the highest overall importance $p_i$.
This steers particle redistribution toward regions that are under-resolved either visually or physically.

\subsection{Differentiable Position Map}
\label{ssec:posmap}
Aligning a rendered shape to its target requires gradients that pull misplaced particles toward the correct silhouette.
Standard image-space losses provide such gradients only indirectly and only where predicted and observed shapes already overlap: rendering losses act through colors and opacities at overlapping pixels, and optical flow supplies inter-frame motion cues that still rely on a roughly aligned reference shape.
Particles that are globally misaligned therefore receive no directional gradient from either loss, a limitation that becomes pronounced under large deformations, where global cues are essential.

We bridge this gap by making each pixel position a differentiable function of the Gaussians that affect that pixel's values, either color or opacity.
By writing each position relative to the covering Gaussian's image-space mean, we introduce the missing gradient path that enables shape supervision.
Let $\mathbf{x}_i^{2D}$ and $\boldsymbol{\Sigma}_i^{2D}$ denote the 2D image-space mean and covariance of Gaussian $i$.
For pixel $\mathbf{p}$ covered by Gaussian $i$, we define the reparameterized position:
\begin{equation}
\label{eq:reparam}
    \tilde{\mathbf{p}}_i = \mathbf{x}_i^{2D} + (\boldsymbol{\Sigma}_i^{2D})^{1/2}\mathbf{z}, \quad \mathbf{z} = \mathrm{sg}\left((\boldsymbol{\Sigma}_i^{2D})^{-1/2}(\mathbf{p} - \mathbf{x}_i^{2D})\right),
\end{equation}
where $\mathrm{sg}(\cdot)$ the stop-gradient operator.
We render a position map using the same alpha-compositing formula as in \cref{eq:3dgs_render}, substituting $\tilde{\mathbf{p}}_i$ for per-Gaussian color:
\begin{equation}
\label{eq:posmap}
    \tilde{\mathbf{p}} = \mathbf{p}\prod_{i}\bigl(1 - \alpha_i\bigr) + \sum_{i} \alpha_i\,\tilde{\mathbf{p}}_i \prod_{j < i}\bigl(1 - \alpha_j\bigr),
\end{equation}
where $\tilde{\mathbf{p}}$ is the output pixel coordinates.
Numerically, $\tilde{\mathbf{p}}$ always equals the actual pixel coordinate $\mathbf{p}$.
This formulation lets pixel-location losses propagate positional gradients to Gaussian parameters.

\subsection{Loss Functions and Optimization}
\label{ssec:optimization}
We introduce the loss terms used across the whole optimization stages.

\noindent \textbf{Image losses ($\mathcal{L}_{\text{color}}$, $\mathcal{L}_{\alpha}$).}
We use standard image losses (L1 + SSIM) for $\mathcal{L}_{\text{color}}$, and L1 loss between the rendered opacity map and the ground-truth alpha mask for $\mathcal{L}_{\alpha}$.

\noindent \textbf{Optical flow loss ($\mathcal{L}_{\text{flow}}$).}
Optical flows can provide which parts should move and in which direction.
Several prior works have similarly used optical flow to supervise motion estimation in differentiable rendering and inverse physics pipelines~\cite{zhu2024motiongs,hong2025physics,liu2025physflow}.
We render predicted flow by splatting per-particle world-space velocities into the image plane using the same Gaussian rasterizer.
Optical flow estimation can be noisy and uncertain, especially when physical dynamics change what is visible between frames.
We therefore use a probability-based optical flow loss~\cite{SEA-RAFT}.
The detailed parameterization is given in the supplementary.
To capture both global displacement and instantaneous motion, we supervise with two complementary optical flow signals: flow from the first frame to the current frame, and flow from the previous frame to the current frame:
\begin{equation}
\label{eq:flow}
    \mathcal{L}_{\text{flow}} = -
    \frac{1}{|\Omega_t|}
    \sum_{j \in \Omega_t}\Big[
        \log p\left(\mathbf{f}_{0 \to t}(\mathbf{p}_j) \mid \hat{\mathbf{f}}_{0 \to t}(\mathbf{p}_j)\right) + \log p\left(\mathbf{f}_{t\text{-}1 \to t}(\mathbf{p}_j) \mid \hat{\mathbf{f}}_{t\text{-}1 \to t}(\mathbf{p}_j)\right)\Big],
\end{equation}
where $\mathbf{p}_j$ is the position of pixel $j$, and $\Omega_t$ is the set of ground-truth foreground pixels at frame $t$.
$\hat{\mathbf{f}}_{i \to j}(\cdot)$ and $\mathbf{f}_{i \to j}(\cdot)$ denote the rendered and estimated optical flow respectively from frame $i$ to frame $j$ at pixel $\mathbf{p}$.
The first term anchors global displacement to the reference frame ($t=0$), while the second term supervises instantaneous velocity.

\noindent \textbf{Silhouette loss ($\mathcal{L}_{\text{sil}}$).}
Using the differentiable position map (\cref{ssec:posmap}), we obtain rendered pixel positions $\tilde{\mathbf{p}}_k$ that are differentiable with respect to Gaussian parameters.
We then treat each foreground region as a weighted point cloud of these positions and align it to the target using optimal transport, which provides global directional gradients even when the predicted and target silhouettes do not overlap.
Let $\Omega_t$ denote the set of target foreground pixels at frame $t$, with per-pixel alpha values $\{a_j\}_{j \in \Omega_t}$, and let $\hat{\Omega}_t$ denote the set of active rendered foreground pixels, with rendered opacity values $\{\tilde{a}_k\}_{k \in \hat{\Omega}_t}$.
Since the number of pixels and their alpha values might differ, we use unbalanced debiased Sinkhorn divergence~\cite{sinkhorn} as a loss:
\begin{equation}
\label{eq:sil_meas}
     \mathcal{L}_{\text{sil}}^t = \mathrm{SD}(\{\tilde{\mathbf{p}}_k,\, \tilde{a}_k\}_{k \in \hat{\Omega}_t},\, \{\mathbf{p}_j,\, a_j\}_{j \in \Omega_t}),
\end{equation}
where $\mathrm{SD}(\cdot, \cdot)$ is the debiased Sinkhorn divergence and $\mathbf{p}_j$ is the GT pixel position.
The resulting gradients indicate how each rendered pixel $\tilde{\mathbf{p}}_k$ should shift spatially and adjust its opacity $\tilde{a}_k$ to match the target silhouette.

\noindent \textbf{Particle distribution regularizer ($\mathcal{L}_{\text{distr}}$).}
With trainable particle positions, the local distribution directly affects inverse physics estimation.
Particles may detach from the body and drift independently, or cluster together and leave voids in their vicinity, both of which degrade the estimation of physical parameters.
To mitigate these failure modes, we regularize the spatial distribution via a K-nearest-neighbor penalty:
\begin{equation}
\label{eq:connectivity}
    \mathcal{L}_{\text{distr}} = \sum_{i=1}^{N}\sum_{j \in \mathcal{N}_{K}(i)} \max\bigl(0,\, \|\mathbf{x}_i - \mathbf{x}_j\| - \tau_{\max} \Delta x\bigr) + \max\bigl(0,\, \tau_{\min} \Delta x - \|\mathbf{x}_i - \mathbf{x}_j\|\bigr),
\end{equation}
where $\mathcal{N}_{K}(i)$ are the $K$ nearest neighbors of particle $i$, $\tau_{\min}$ and $\tau_{\max}$ are lower and upper bounds in the unit of simulation cell size $\Delta x$.
The first term pulls a particle toward its neighbors when they drift too far apart, while the second pushes neighbors apart when they become overly clustered.
Both terms vanish in well-distributed regions, avoiding over-regularization of the particle distribution and the underlying geometry.
Since each Gaussian can undergo a large deformation during simulation, we apply this regularizer only at the initial configuration (frame $0$), before any deformation occurs.

\noindent \textbf{Optimization strategy.}
Starting from initial 3D geometry obtained from an image-to-3D model~\cite{sam3dteam2025sam3d3dfyimages}, our optimization proceeds in two stages.
In the first stage, we jointly optimize all physics parameters (material parameters and initial velocity), Gaussian parameters (positions $\mathbf{x}_i$, opacity $o_i$, and covariance $\boldsymbol{\Sigma}_i$), and global scale $s$ (\cref{ssec:scale}) via differentiable MPM simulation, while not optimizing colors $\mathbf{c}_i$.
Following GIC~\cite{GIC}, we focus on physical dynamics in this stage and refine appearance later.
Within each iteration, after backpropagating through the simulation, we take a small number of alpha-map loss steps without re-running the simulation.
This keeps the visual representation consistent with the evolving geometry.
We use all four losses as follows:
\begin{equation}
\label{eq:total_loss_1}
    \mathcal{L} = \sum_{t=0}^{T}\left[ \mathcal{L}_{\alpha}^t + \mathcal{L}_{\text{sil}}^t\right] + \sum_{t=1}^{T}\mathcal{L}_{\text{flow}}^t + \mathcal{L}_{\text{distr}}.
\end{equation}
In the second stage, we only refine appearance, using both image $\mathcal{L}_{\text{color}}$ and alpha $\mathcal{L}_{\alpha}$ losses.
Details are in the supplementary.

\begin{figure*}[!tp]
    {\captionsetup{position=top}
\centering
    \captionof{table}{
        \textbf{Evaluation on the Vid2Sim dataset.}
        We report rendering quality metrics for future frame prediction and material parameter estimation errors.
        Monocular results are median [Q1, Q3].
        Multi-view results, denoted by~$*$, are taken from Vid2Sim~\cite{vid2sim} and reported as single values.
        Bold marks the best monocular method in each column, excluding \textit{Oracle Midpoint}.
    }
    \label{tab:main_table_v2s}
    \resizebox{\linewidth}{!}{%
    \begin{tabular}{lcccccc}
    \toprule
     & & \multicolumn{3}{c}{\textbf{Future Frame Rendering}} & \multicolumn{2}{c}{\textbf{Physical Parameter Estimation}} \\
    \cmidrule(lr){3-5} \cmidrule(lr){6-7}
    & \textbf{views} & \textbf{PSNR $\uparrow$}  
    & \textbf{LPIPS $\downarrow$} 
    & \textbf{SSIM $\uparrow$} 
    & \textbf{MAE $\log E$ $\downarrow$} 
    & \textbf{MAE $\nu$ $\downarrow$} \\
    \midrule
    Vid2Sim* & 12 & 25.07 & - & 0.95 & 0.51 & 0.06 \\
    \midrule
    \multicolumn{2}{l}{\textit{Oracle Midpoint}} & & & & 0.57 [0.25, 0.77] & 0.03 [0.02, 0.07] \\
    PAC-NeRF & 1 & 17.9 [16.6, 18.8] & 0.15 [0.14, 0.19] & 0.91 [0.89, 0.93] & 2.07 [0.97, 2.83] & 0.14 [0.08, 0.36] \\
    SpringGaus & 1 & 18.9 [16.4, 20.6] & 0.16 [0.10, 0.20] & 0.91 [0.89, 0.94] & -- & -- \\ 
    GIC & 1 & 18.7 [17.5, 20.3] & 0.16 [0.14, 0.18] & 0.92 [0.90, 0.93] & 0.48 [0.25, 0.77] & \textbf{0.07} [0.03, 0.10] \\
    \textbf{Ours} & 1 & \textbf{22.4} [20.5, 23.6] & \textbf{0.09} [0.07, 0.13] & \textbf{0.94} [0.92, 0.95] & \textbf{0.45} [0.21, 0.80] & 0.10 [0.06, 0.13] \\
    \bottomrule
    \end{tabular}
    }

}
    \vspace{0.2em}
    \centering
    \setlength{\tabcolsep}{1pt} 
    \setlength{\fboxsep}{0pt}     
    \setlength{\fboxrule}{0.6pt}  

    \begin{tabular*}{\textwidth}{@{\extracolsep{\fill}}c c c c c c c@{}}
     & 
    \multicolumn{3}{c}{%
      \begin{tikzpicture}
        \draw[->, thick] (0,0) -- (0.45\textwidth,0) node[midway, above]{\footnotesize\textbf{Time}};
      \end{tikzpicture}
    } &
    \multicolumn{3}{c}{%
      \begin{tikzpicture}
        \draw[->, thick] (0,0) -- (0.45\textwidth,0) node[midway, above]{\footnotesize\textbf{Time}};
      \end{tikzpicture}
    } \\


    
    \rotatebox{90}{\makebox[0.15\linewidth]{\footnotesize\textbf{G.T}}} &
    \fbox{\includegraphics[width=0.15\linewidth]{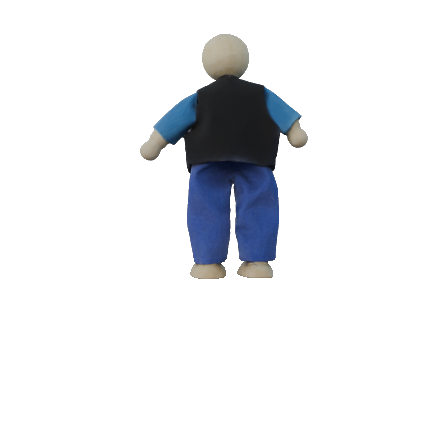}} &
    \fbox{\includegraphics[width=0.15\linewidth]{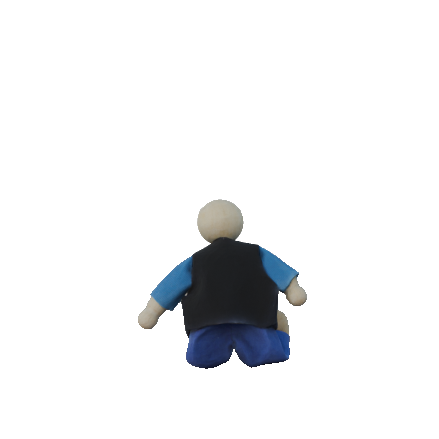}} &
    \fbox{\includegraphics[width=0.15\linewidth]{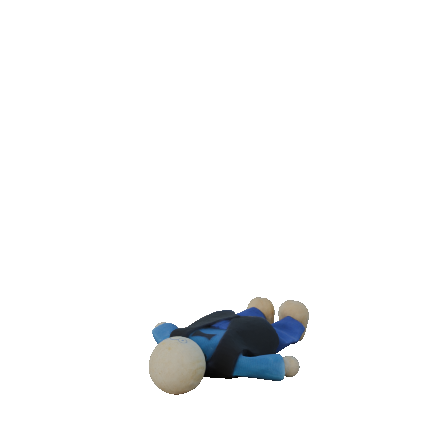}} &
    \fbox{\includegraphics[width=0.15\linewidth]{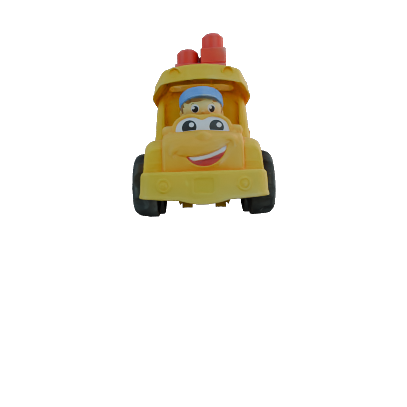}} &
    \fbox{\includegraphics[width=0.15\linewidth]{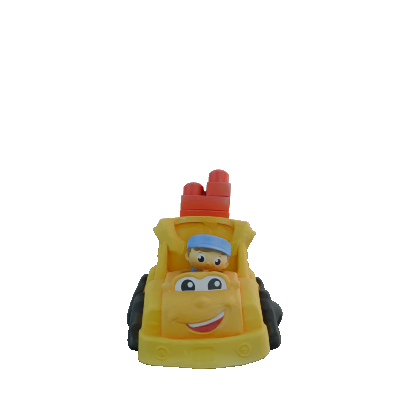}} &
    \fbox{\includegraphics[width=0.15\linewidth]{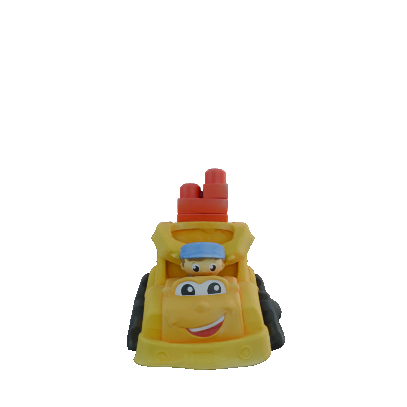}}
    \\
    
    
    \rotatebox{90}{\makebox[0.15\linewidth]{\footnotesize\textbf{Ours}}} &
    \fbox{\includegraphics[width=0.15\linewidth]{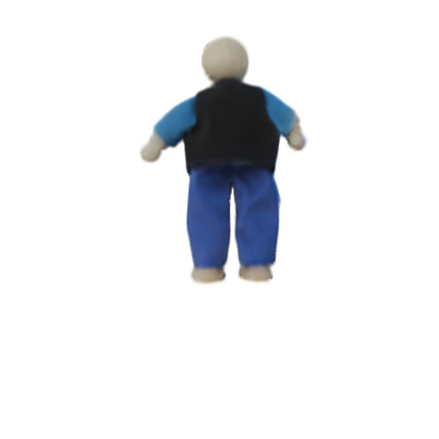}} &
    \fbox{\includegraphics[width=0.15\linewidth]{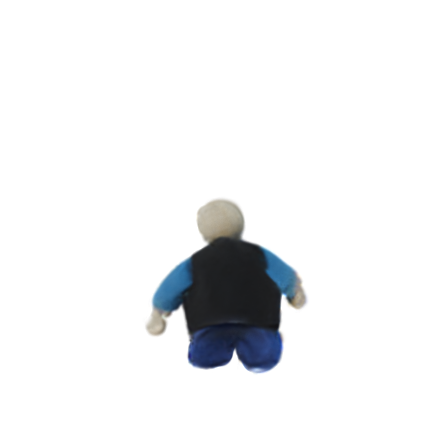}} &
    \fbox{\includegraphics[width=0.15\linewidth]{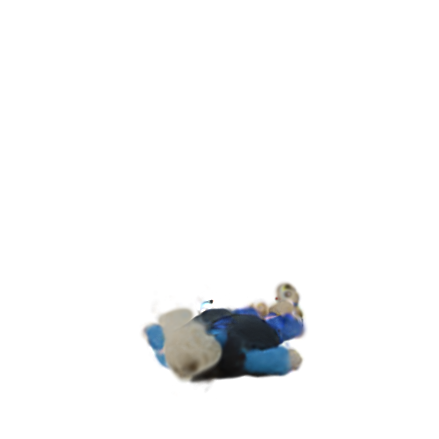}} &
    \fbox{\includegraphics[width=0.15\linewidth]{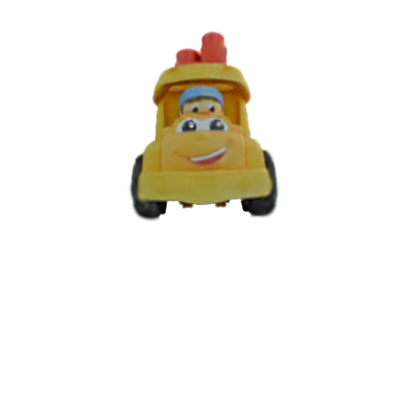}} &
    \fbox{\includegraphics[width=0.15\linewidth]{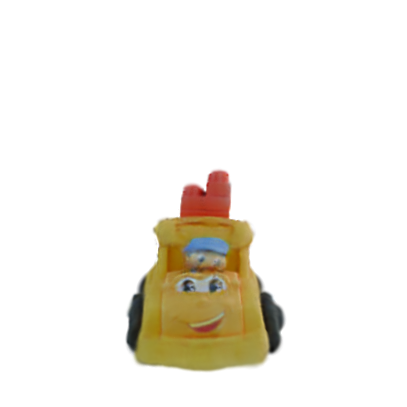}} &
    \fbox{\includegraphics[width=0.15\linewidth]{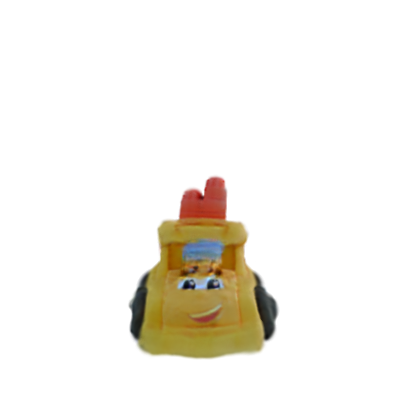}}
    \\
    
    \rotatebox{90}{\makebox[0.15\linewidth]{\footnotesize\textbf{GIC}}} &
    \fbox{\includegraphics[width=0.15\linewidth]{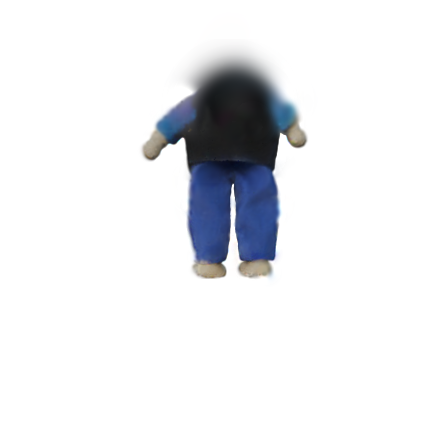}} &
    \fbox{\includegraphics[width=0.15\linewidth]{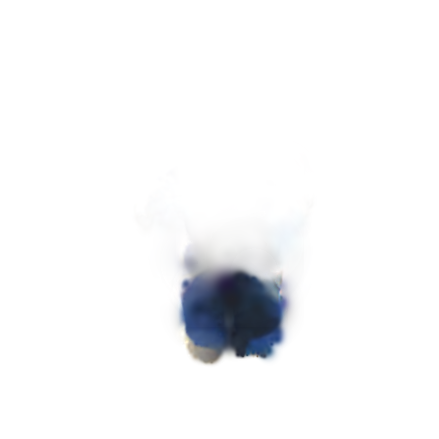}} &
    \fbox{\includegraphics[width=0.15\linewidth]{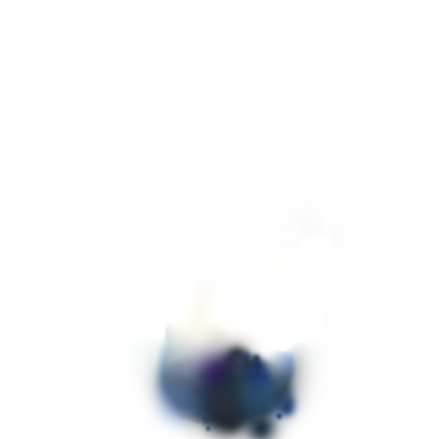}} &
    \fbox{\includegraphics[width=0.15\linewidth]{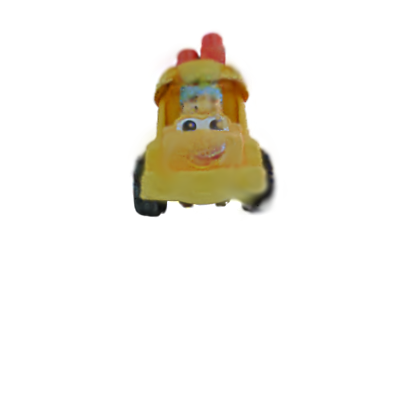}} &
    \fbox{\includegraphics[width=0.15\linewidth]{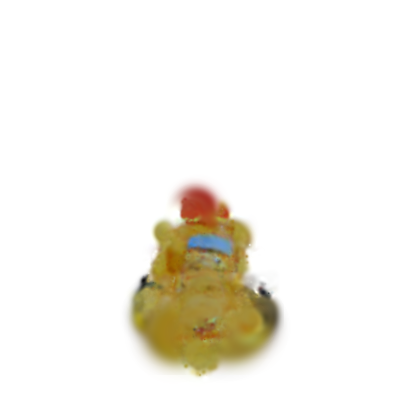}} &
    \fbox{\includegraphics[width=0.15\linewidth]{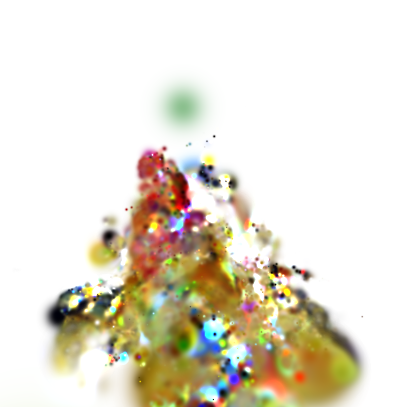}}
    \\
    
    \rotatebox{90}{\makebox[0.15\linewidth]{\footnotesize\textbf{SpringGaus}}} &
    \fbox{\includegraphics[width=0.15\linewidth]{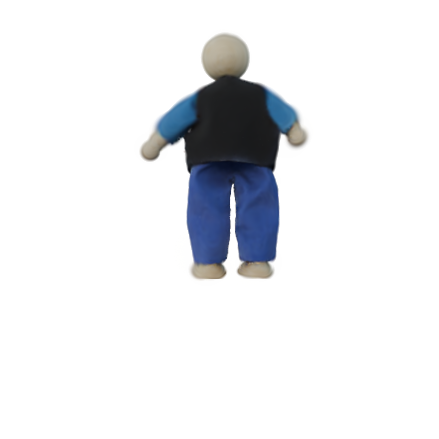}} &
    \fbox{\includegraphics[width=0.15\linewidth]{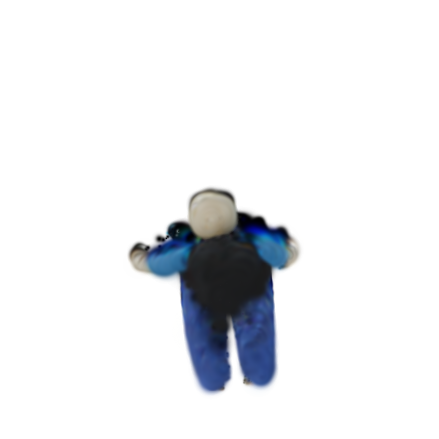}} &
    \fbox{\includegraphics[width=0.15\linewidth]{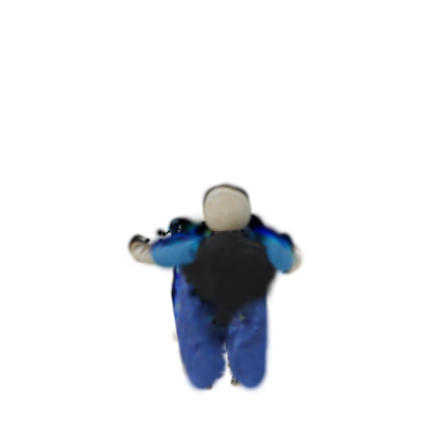}} &
    \fbox{\includegraphics[width=0.15\linewidth]{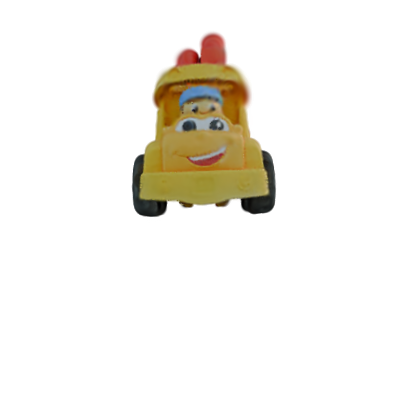}} &
    \fbox{\includegraphics[width=0.15\linewidth]{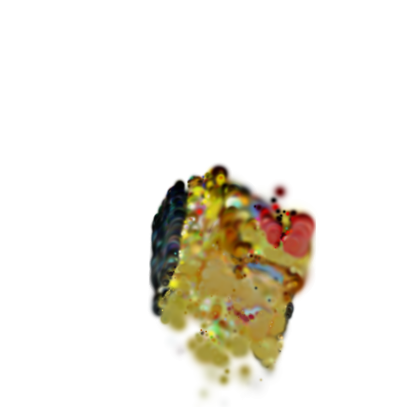}} &
    \fbox{\includegraphics[width=0.15\linewidth]{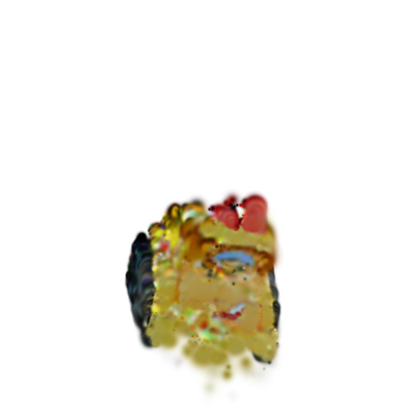}}
    \\

    \end{tabular*}
    \caption{
        \textbf{Qualitative comparison on Vid2Sim dataset.}
        We compare the rendered image trajectory over time for SpringGaus~\cite{SpringGaus}, GIC \cite{GIC}, and our method.
    }
    \label{fig:main_figure_Vid2Sim}
\end{figure*}
\begin{figure*}[!tp]
    {\captionsetup{position=top}
    \captionof{table}{
        \textbf{Evaluation on our synthetic dataset.}
        We evaluate 3D trajectory accuracy and rendering quality on future frames.
        We also evaluate physical parameter estimation.
        Values are medians.
        Due to space, the full results with median [Q1, Q3] are in the supplementary material.
        Bold marks the best method in each column, excluding \textit{Oracle Midpoint}.
    }
    \label{tab:main_table_Ours}
    \centering
    \resizebox{\linewidth}{!}{%
        \begin{tabular}{lccccccccc}
        \toprule
        & \multicolumn{2}{c}{\textbf{3D Trajectory}} & \multicolumn{3}{c}{\textbf{Future Frame Rendering}} &\multicolumn{4}{c}{\textbf{Physical Parameter Estimation}} \\

        \cmidrule(lr){2-3} \cmidrule(lr){4-6} \cmidrule(lr){7-10}
        & \textbf{CD $\downarrow$} & \textbf{EMD $\downarrow$}
        & \textbf{PSNR $\uparrow$}  & \textbf{LPIPS $\downarrow$} & \textbf{SSIM $\uparrow$}
         & \textbf{MAE $\mathbf{v}_0$ $\downarrow$}
        & \textbf{MAE $\log E$ $\downarrow$} & \textbf{MAE $\nu$ $\downarrow$}
        & \textbf{MAE $\log \sigma_y$ $\downarrow$} \\
        \midrule

        \multicolumn{9}{l}{\textbf{Elastic}} \\
        \midrule
        \multicolumn{2}{l}{\textit{Oracle Midpoint}} & & & & & 0.15 & 1.11 & 0.20 & - \\
        PAC-NeRF & 438 & 0.74 & 17.4 & 0.19 & 0.92 & \textbf{0.04} & 1.33 & \textbf{0.06} & - \\
        SpringGaus & 693 & 0.97 & 16.7 & 0.24 & 0.91 & 0.16 & - & - & - \\
        GIC & 421 & 0.75 & 18.5 & 0.18 & 0.92 & 0.09 & 0.66 & 0.21 & - \\
        \textbf{Ours} & \ \ \textbf{24} & \textbf{0.26} & \textbf{21.1} & \textbf{0.12} & \textbf{0.93} & 0.09 & \textbf{0.31} & 0.07 & - \\

        \midrule

        \multicolumn{9}{l}{\textbf{Plasticine}} \\
        \midrule
        \multicolumn{2}{l}{\textit{Oracle Midpoint}} & & & & & 0.22 & 0.32 & 0.01 & 0.81 \\
        PAC-NeRF & 247 & 0.62 & 19.8 & 0.14 & 0.94 & 0.14 & 2.55 & 0.17 & 0.57 \\
        SpringGaus & 450 & 0.67 & 17.3 & 0.20 & 0.92 & 0.13 & - & - & - \\
        GIC & 333 & 0.65 & 22.0 & 0.14 & 0.94 & \textbf{0.12} & 0.95 & 0.07 & 0.34 \\
        \textbf{Ours} & \ \ \textbf{19} & \textbf{0.24} & \textbf{23.4} & \textbf{0.09} & \textbf{0.95} & 0.14 & \textbf{0.40} & \textbf{0.05} & \textbf{0.33} \\
        \bottomrule
        \end{tabular}%
    }
}
    \vspace{0.2em}
    \centering
    \setlength{\tabcolsep}{1pt} 
    \setlength{\fboxsep}{0pt}     
    \setlength{\fboxrule}{0.6pt}  

    \begin{tabular*}{\textwidth}{@{\extracolsep{\fill}}c c c c c c c@{}}
     & 
    \multicolumn{3}{c}{%
      \begin{tikzpicture}
        \draw[->, thick] (0,0) -- (0.45\textwidth,0) node[midway, above]{\footnotesize\textbf{Time}};
      \end{tikzpicture}
    } &
    \multicolumn{3}{c}{%
      \begin{tikzpicture}
        \draw[->, thick] (0,0) -- (0.45\textwidth,0) node[midway, above]{\footnotesize\textbf{Time}};
      \end{tikzpicture}
    } \\
    
    
    
    \rotatebox{90}{\makebox[0.15\linewidth]{\footnotesize\textbf{G.T}}} &
    \fbox{\includegraphics[width=0.15\linewidth]{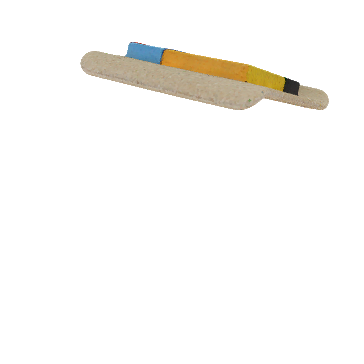}} &
    \fbox{\includegraphics[width=0.15\linewidth]{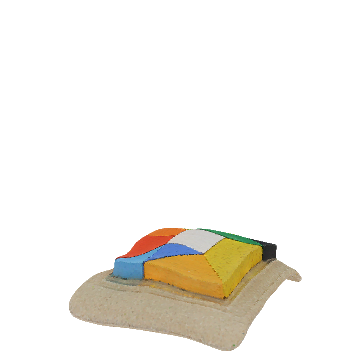}} &
    \fbox{\includegraphics[width=0.15\linewidth]{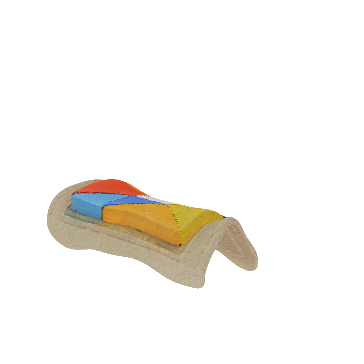}} &
    \fbox{\includegraphics[width=0.15\linewidth]{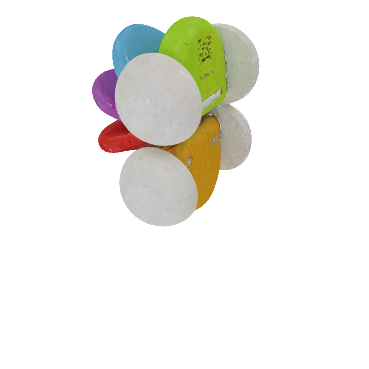}} &
    \fbox{\includegraphics[width=0.15\linewidth]{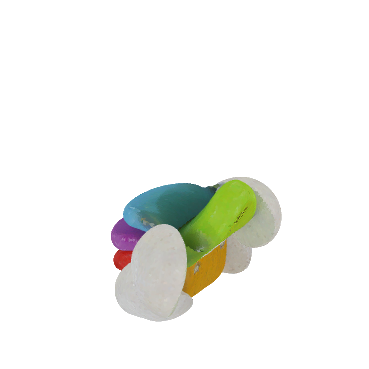}} &
    \fbox{\includegraphics[width=0.15\linewidth]{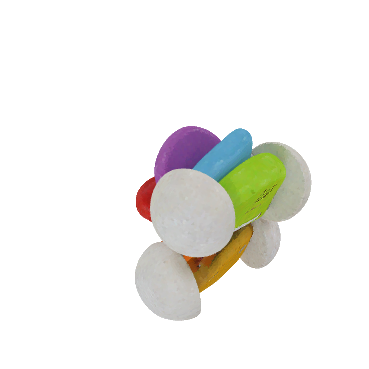}}
    \\
    
    
    \rotatebox{90}{\makebox[0.15\linewidth]{\footnotesize\textbf{Ours}}} &
    \fbox{\includegraphics[width=0.15\linewidth]{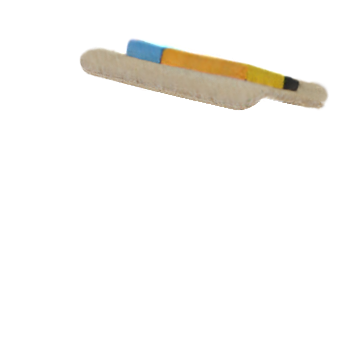}} &
    \fbox{\includegraphics[width=0.15\linewidth]{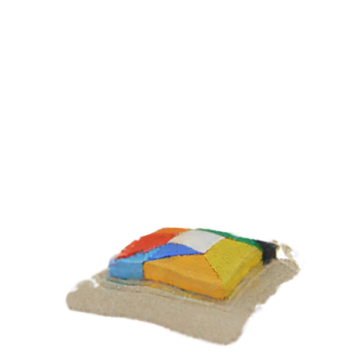}} &
    \fbox{\includegraphics[width=0.15\linewidth]{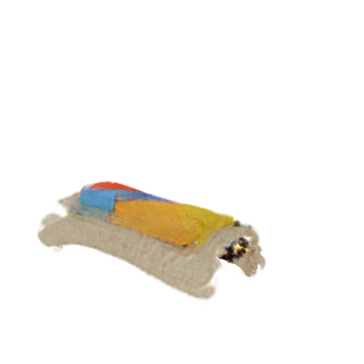}} &
    \fbox{\includegraphics[width=0.15\linewidth]{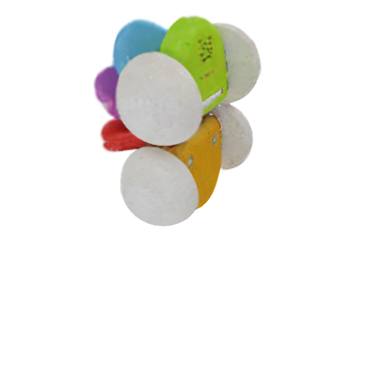}} &
    \fbox{\includegraphics[width=0.15\linewidth]{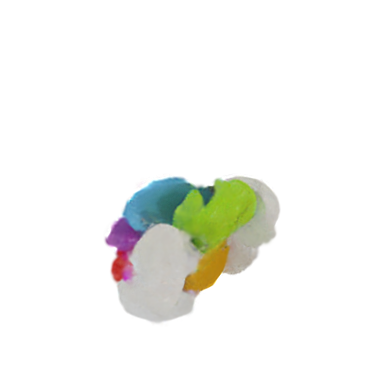}} &
    \fbox{\includegraphics[width=0.15\linewidth]{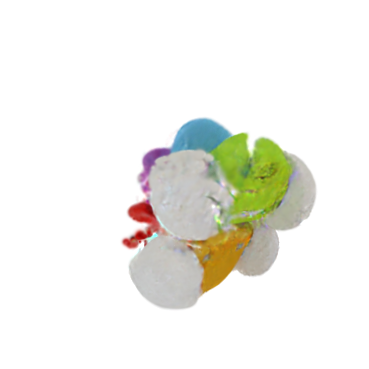}}
    \\
    
    \rotatebox{90}{\makebox[0.15\linewidth]{\footnotesize\textbf{GIC}}} &
    \fbox{\includegraphics[width=0.15\linewidth]{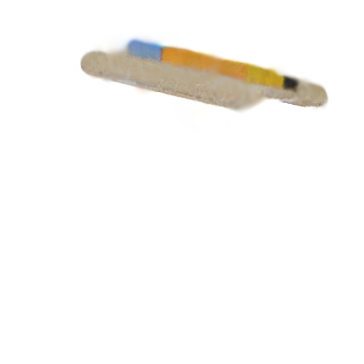}} &
    \fbox{\includegraphics[width=0.15\linewidth]{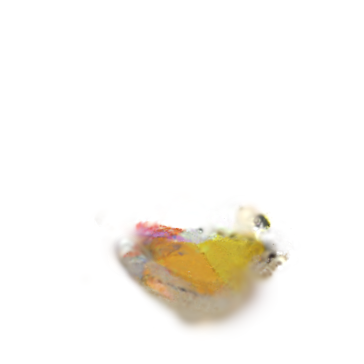}} &
    \fbox{\includegraphics[width=0.15\linewidth]{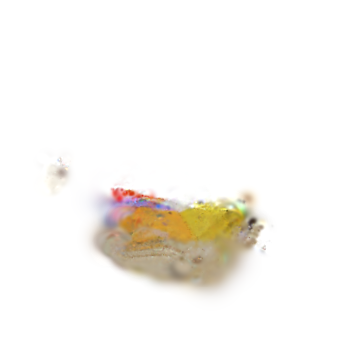}} &
    \fbox{\includegraphics[width=0.15\linewidth]{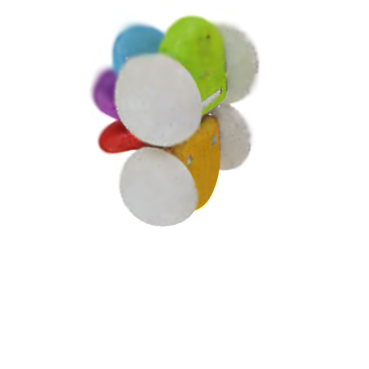}} &
    \fbox{\includegraphics[width=0.15\linewidth]{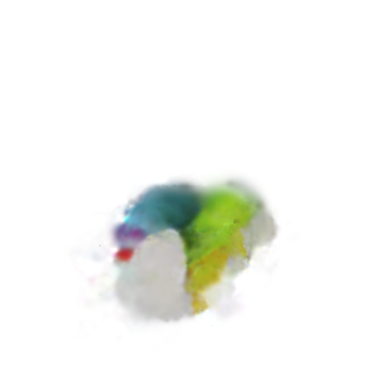}} &
    \fbox{\includegraphics[width=0.15\linewidth]{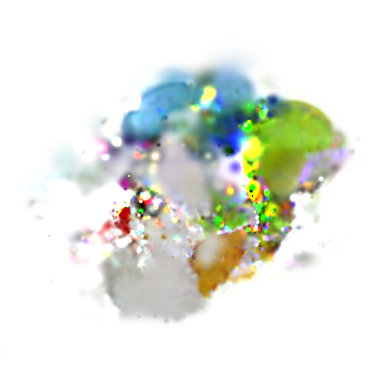}}
    \\
    
    \rotatebox{90}{\makebox[0.15\linewidth]{\footnotesize\textbf{SpringGaus}}} &
    \fbox{\includegraphics[width=0.15\linewidth]{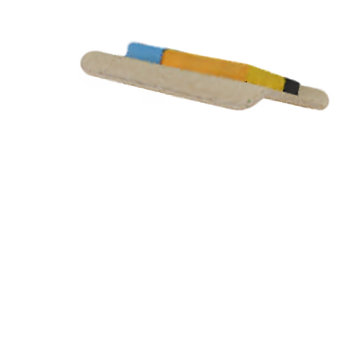}} &
    \fbox{\includegraphics[width=0.15\linewidth]{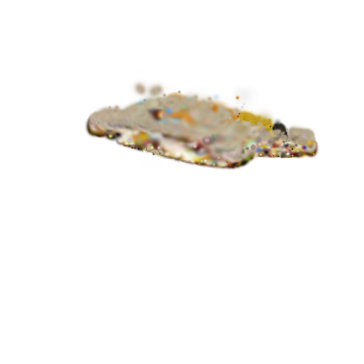}} &
    \fbox{\includegraphics[width=0.15\linewidth]{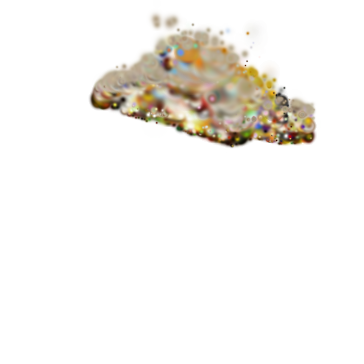}} &
    \fbox{\includegraphics[width=0.15\linewidth]{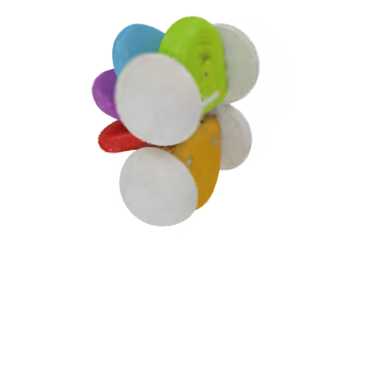}} &
    \fbox{\includegraphics[width=0.15\linewidth]{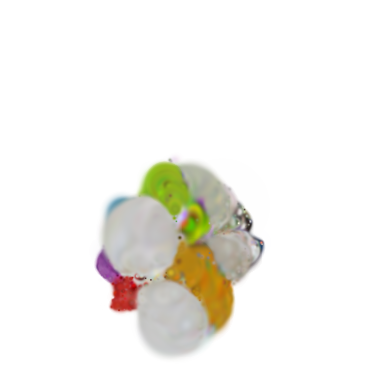}} &
    \fbox{\includegraphics[width=0.15\linewidth]{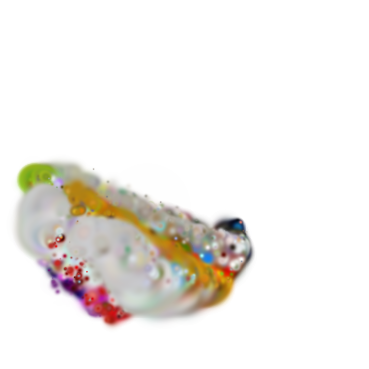}}
    \\
    
    \end{tabular*}
    \caption{
        \textbf{Qualitative comparison on our synthetic dataset.}
        We compare the rendered image trajectory over time for SpringGaus~\cite{SpringGaus}, GIC \cite{GIC}, and our method.
        (Left: elastic. Right: plasticine.)
    }
    \label{fig:main_figure_Ours}
    \vspace{-10pt}
\end{figure*}
\begin{figure*}[!tp]
    {\captionsetup{position=top}
  \captionof{table}{
    \textbf{Evaluation on real captures from SpringGaus~\cite{SpringGaus}.}
    Values are median [Q1, Q3].
  }
  \label{tab:real}
  \centering
  \resizebox{0.8\linewidth}{!}{%
  \begin{tabular}{lcccc}
    \toprule
    & \textbf{PSNR $\uparrow$}  & \textbf{LPIPS $\downarrow$} & \textbf{SSIM $\uparrow$} \\
    \midrule
    SpringGaus & 28.94 [26.28, 29.91] & 0.0128 [0.0109, 0.0183] & 0.9952 [0.9934, 0.9958] \\
    GIC & 25.89 [25.16, 26.38] & 0.0255 [0.0234, 0.0282] & 0.9922 [0.9914, 0.9928] \\
    \textbf{Ours} & \textbf{32.33} [31.46, 33.38] & \textbf{0.0095} [0.0084, 0.0102] & \textbf{0.9957} [0.9950, 0.9962] \\
    \bottomrule
  \end{tabular}%
  }

}
    \vspace{0.2em}
    \centering
    \setlength{\tabcolsep}{1pt}
    \setlength{\fboxsep}{0pt}     
    \setlength{\fboxrule}{0.6pt}  

    \begin{tabular*}{\textwidth}{@{\extracolsep{\fill}}c c c c c c c@{}}
     &
    \multicolumn{3}{c}{%
      \begin{tikzpicture}
        \draw[->, thick] (0,0) -- (0.45\textwidth,0) node[midway, above]{\footnotesize\textbf{Time}};
      \end{tikzpicture}
    } &
    \multicolumn{3}{c}{%
      \begin{tikzpicture}
        \draw[->, thick] (0,0) -- (0.45\textwidth,0) node[midway, above]{\footnotesize\textbf{Time}};
      \end{tikzpicture}
    } \\



    \rotatebox{90}{\makebox[0.15\linewidth]{\footnotesize\textbf{G.T}}} &
    \fbox{\includegraphics[width=0.15\linewidth]{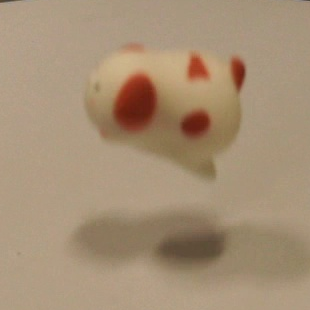}} &
    \fbox{\includegraphics[width=0.15\linewidth]{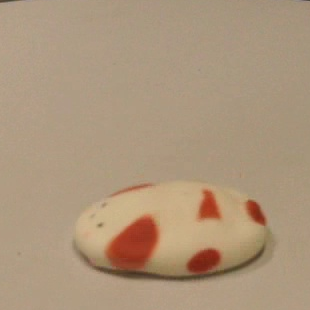}} &
    \fbox{\includegraphics[width=0.15\linewidth]{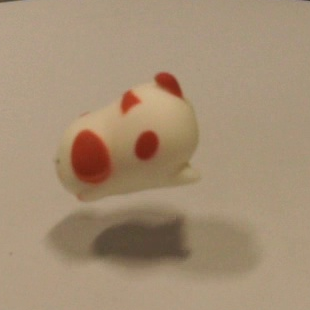}} &
    \fbox{\includegraphics[width=0.15\linewidth]{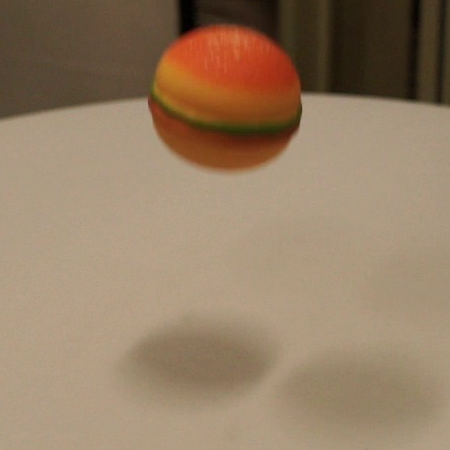}} &
    \fbox{\includegraphics[width=0.15\linewidth]{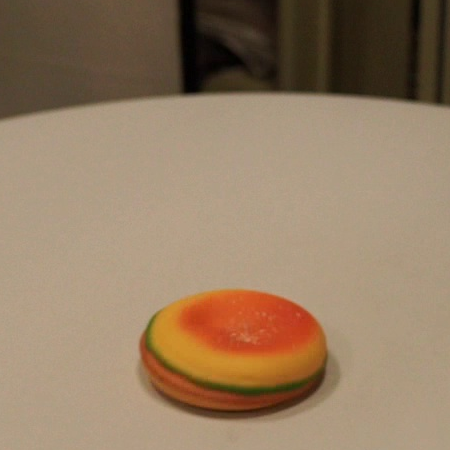}} &
    \fbox{\includegraphics[width=0.15\linewidth]{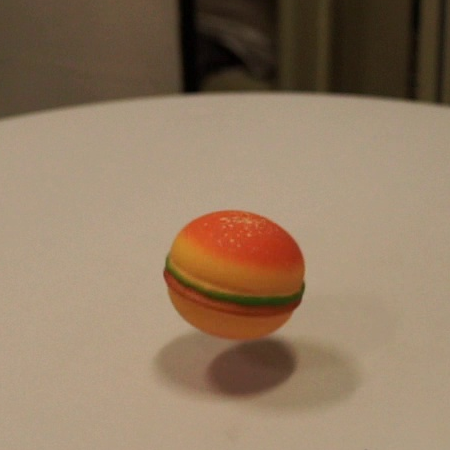}}
    \\

    \rotatebox{90}{\makebox[0.15\linewidth]{\footnotesize\textbf{Ours}}} &
    \fbox{\includegraphics[width=0.15\linewidth]{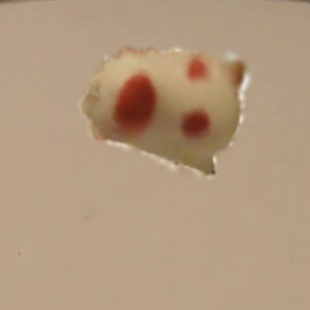}} &
    \fbox{\includegraphics[width=0.15\linewidth]{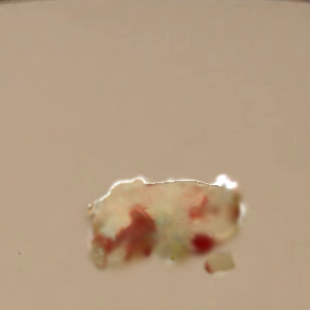}} &
    \fbox{\includegraphics[width=0.15\linewidth]{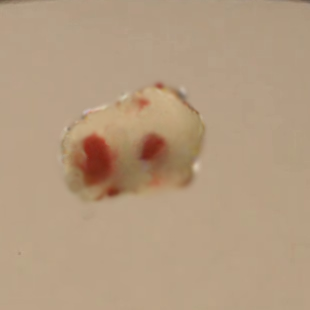}} &
    \fbox{\includegraphics[width=0.15\linewidth]{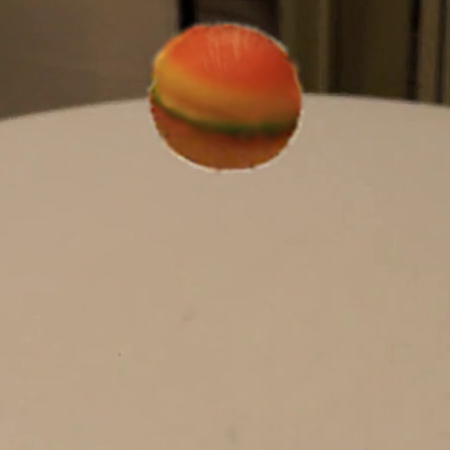}} &
    \fbox{\includegraphics[width=0.15\linewidth]{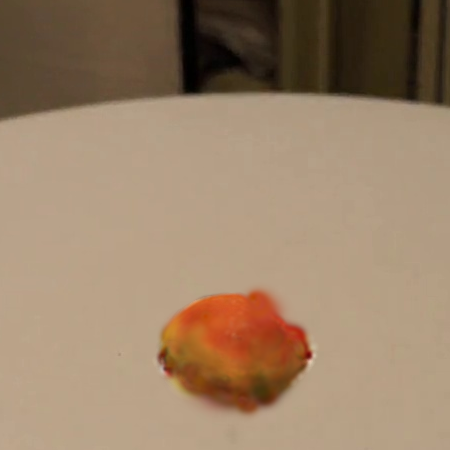}} &
    \fbox{\includegraphics[width=0.15\linewidth]{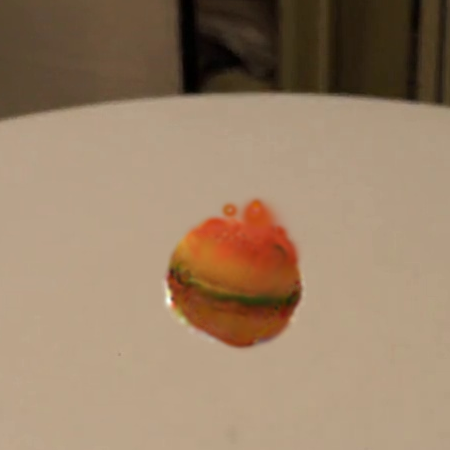}}
    \\

    \rotatebox{90}{\makebox[0.15\linewidth]{\footnotesize\textbf{GIC}}} &
    \fbox{\includegraphics[width=0.15\linewidth]{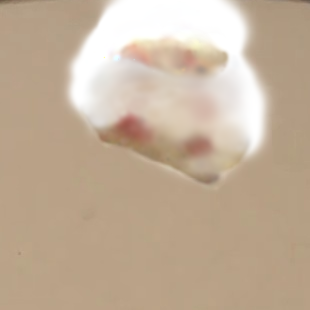}} &
    \fbox{\includegraphics[width=0.15\linewidth]{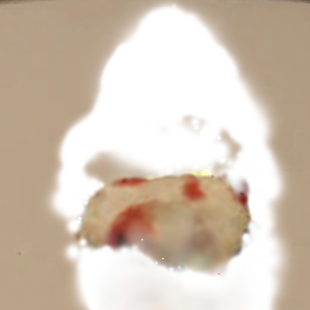}} &
    \fbox{\includegraphics[width=0.15\linewidth]{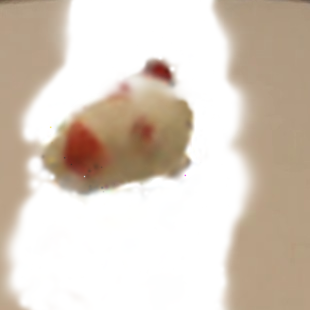}} &
    \fbox{\includegraphics[width=0.15\linewidth]{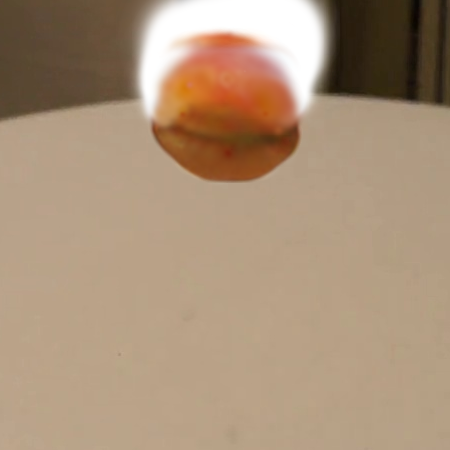}} &
    \fbox{\includegraphics[width=0.15\linewidth]{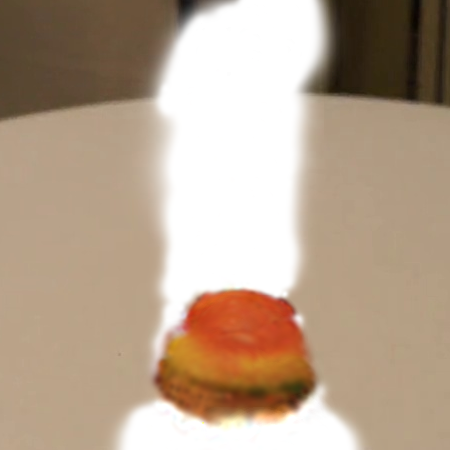}} &
    \fbox{\includegraphics[width=0.15\linewidth]{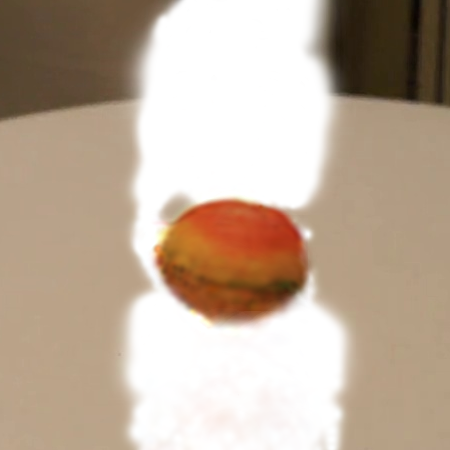}}
    \\

    \rotatebox{90}{\makebox[0.15\linewidth]{\footnotesize\textbf{SpringGaus}}} &
    \fbox{\includegraphics[width=0.15\linewidth]{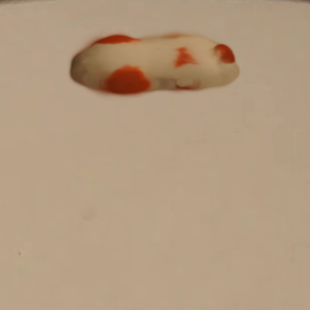}} &
    \fbox{\includegraphics[width=0.15\linewidth]{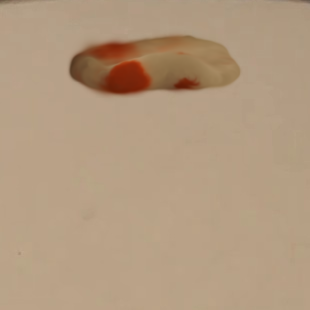}} &
    \fbox{\includegraphics[width=0.15\linewidth]{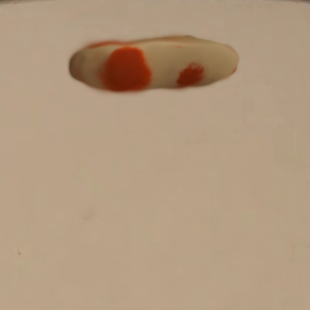}} &
    \fbox{\includegraphics[width=0.15\linewidth]{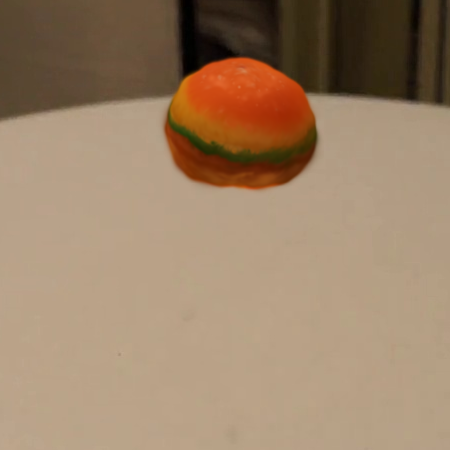}} &
    \fbox{\includegraphics[width=0.15\linewidth]{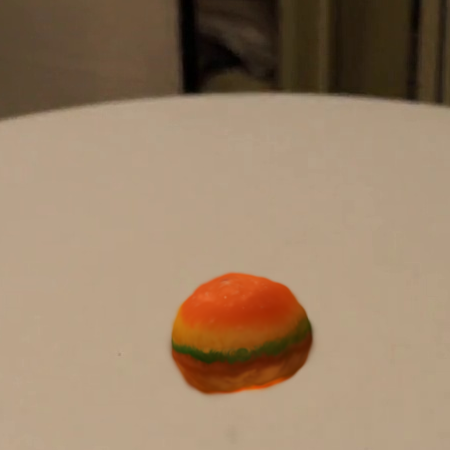}} &
    \fbox{\includegraphics[width=0.15\linewidth]{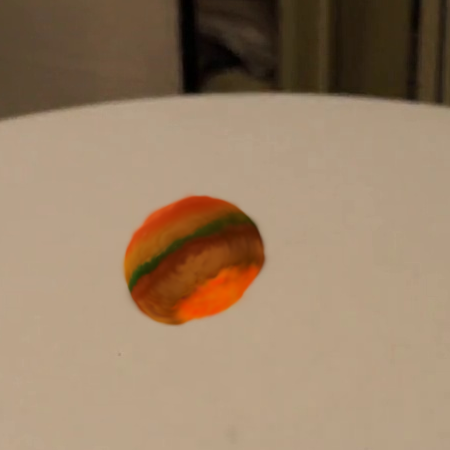}}
    \\

    \end{tabular*}
    \caption{
        \textbf{Qualitative comparison on SpringGaus real-world dataset.}
        We compare the rendered image trajectory over time for SpringGaus~\cite{SpringGaus}, GIC \cite{GIC}, and our method.
    }
    \label{fig:main_figure_real}

\end{figure*}

\section{Experiments}
\label{sec:experiments}

\subsection{Experimental Settings}
\label{ssec:exp_settings}

\noindent\textbf{Baselines.}
We compare our method against existing inverse physics methods: PAC-NeRF~\cite{li2023pacnerf}, SpringGaus~\cite{SpringGaus}, GIC~\cite{GIC}, and Vid2Sim~\cite{vid2sim}.
Since Vid2Sim requires multiple views and cannot operate on monocular video input, we report only its multi-view results, taken from the original paper.
We additionally report \textit{Oracle Midpoint}, which ignores the input video and always predicts the midpoint of each parameter's sampling range.
Since it has access to the sampling range, which is unavailable to the other methods, it serves only as a reference.
Outperforming it indicates that a method recovers information from the video beyond what the range alone provides.

\noindent\textbf{Datasets.}
We evaluate on two synthetic datasets and one real-world dataset.
First, we use the Vid2Sim~\cite{vid2sim} dataset, which contains 12 elastic objects with no initial velocity.
We run each scene independently from 5 different cameras, treating each camera as a separate monocular input (60 runs per method in total).
Second, we introduce a new synthetic dataset built from Google Scanned Objects (GSO, CC-BY 4.0)~\cite{GSO}, comprising 5 elastic (Neo-Hookean) and 5 plasticine objects.
Each scene has randomly sampled material parameters, initial velocity, object size, and camera distance.
As with Vid2Sim, each scene is run from 5 different cameras, giving 25 runs per material.
For real-world evaluation, we use the real captures released with SpringGaus~\cite{SpringGaus} (5 scenes with 3 cameras each).
Since the real captures have no ground-truth material parameters, we report only the rendering quality of future predictions (\cref{tab:real}). 

\noindent\textbf{Metrics.}
We report three groups of metrics.
For future prediction, we use PSNR, SSIM, and LPIPS, evaluated from the same camera viewpoint.
For geometric accuracy, we use Chamfer Distance (CD) and Earth Mover's Distance (EMD).
CD is computed as the bidirectional Chamfer distance between the predicted Gaussian positions and the ground-truth point cloud.
Both CD and EMD are averaged over the future frames.
For physical parameter estimation, we report the mean absolute error (MAE), following the standard evaluation protocol~\cite{li2023pacnerf,SpringGaus,GIC}.
Since Young's modulus $E$ and density cannot be identified independently from motion, all methods use a known density, following prior work.
Therefore, the MAE of $\log E$ should be interpreted as accuracy under a known density.
Because several metrics (CD in particular) are strongly skewed, we report the median and interquartile range.
Due to space, \cref{tab:main_table_Ours} reports medians only, and its full results are in the supplementary material.

\subsection{Comparison}
\label{ssec:comparison}

\noindent\textbf{Vid2Sim benchmark.}
\cref{tab:main_table_v2s} and \cref{fig:main_figure_Vid2Sim} show results on Vid2Sim.
Among monocular methods, ours achieves the best rendering quality on all three metrics and the lowest MAE $\log E$.
Notably, our MAE $\log E$ is on par with multi-view Vid2Sim using 12 cameras (0.45 vs.\ 0.51).
For $\nu$, GIC achieves a slightly lower MAE than ours, and no monocular method outperforms \textit{Oracle Midpoint}.

\noindent\textbf{Our synthetic benchmark.}
\cref{tab:main_table_Ours} and \cref{fig:main_figure_Ours} provide results on our synthetic dataset.
Our method achieves the lowest CD and EMD for both material types, indicating more accurate geometry recovery, and the best rendering quality on all three metrics.
We attribute these improvements to resolving scale ambiguity and jointly refining geometry with physical parameters.
As shown in \cref{fig:main_figure_Ours}, our reconstructions follow the observed deformation and maintain realistic contact with the ground, whereas the baselines often produce incorrectly scaled or distorted shapes.

For material parameters, our method achieves the lowest MAE on Young's modulus for both materials, and on Poisson's ratio and yield stress for plasticine.
However, yield stress is the only plasticine material parameter for which any method outperforms \textit{Oracle Midpoint}.
SpringGaus is excluded from the material parameter comparison because it uses a spring-mass simulator whose parameters are not directly comparable to those of the other methods.
Initial-velocity errors are similar across methods, except that PAC-NeRF achieves notably lower error on elastic objects.

\noindent\textbf{Real-world captures.}
\cref{tab:real} and \cref{fig:main_figure_real} show results on the real captures from SpringGaus~\cite{SpringGaus}.
Our method predicts future frames more accurately than the baselines on all three rendering metrics, showing that it also works on real captures and not only on synthetic data.

\subsection{Ablation Study}
\label{ssec:ablation_study}

\begin{table}[t]
    \vspace{-15pt}
    \centering
    \captionof{table}{
        \textbf{Ablation study on loss components.}
        Values are median [Q1, Q3].
    }
    \label{tab:loss_ablation}
    \resizebox{0.9\linewidth}{!}{
    \begin{tabular}{ccccccccc}
    \toprule
    $\mathcal{L}_{\alpha}$ & $\mathcal{L}_{\text{distr}}$ & $\mathcal{L}_{\text{flow}}$ & $\mathcal{L}_{\text{sil}}$ & CD $\downarrow$ & PSNR $\uparrow$ & MAE $\log E$ $\downarrow$ & MAE $\log \sigma_y$ $\downarrow$ \\
    \midrule
    \checkmark & -- & -- & -- & 78 [33, 294] & 21.8 [20.3, 23.9] & 0.34 [0.13, 0.82] & 0.45 [0.30, 0.98] \\
    \checkmark & \checkmark & -- & -- & 64 [14, 183] & 22.2 [19.8, 28.1] & \textbf{0.23} [0.13, 1.02] & 0.27 [0.11, 0.62] \\
    \checkmark & \checkmark & \checkmark & -- & 71 [32, 153] & 22.9 [20.8, 24.9] & 0.50 [0.17, 1.18] & \textbf{0.21} [0.16, 0.53] \\
    \checkmark & \checkmark & -- & \checkmark & 26 [11, \ \ 72] & 22.9 [21.4, 24.4] & 0.36 [0.17, 0.64] & 0.52 [0.28, 0.85] \\
    \checkmark & \checkmark & \checkmark & \checkmark & \textbf{19} [\ \ 7, \ \ 72] & \textbf{23.4} [22.4, 25.4] & 0.40 [0.14, 0.81] & 0.33 [0.19, 0.50] \\
    \bottomrule
    \end{tabular}
    }
\end{table}

\noindent\textbf{Loss Functions.}
\cref{tab:loss_ablation} ablates the contribution of each loss from \cref{ssec:optimization} on the plasticine subset of our synthetic dataset.
Adding the distribution regularizer to the baseline reduces CD and lowers both material parameter errors, consistent with its role in preventing disconnected particles from simulating independently and distorting future predictions.
However, without positional guidance from optical flow or silhouette supervision, CD remains high.
Optical flow supervision alone does not reduce CD.
In contrast, silhouette supervision alone (\cref{ssec:posmap}) yields the largest improvement in trajectory and geometry (CD from 64 to 26), reflecting its role as a global shape-alignment signal.
Combining both losses achieves the lowest CD and the highest PSNR, indicating that flow and silhouette supervision are complementary: flow anchors the physical trajectory, while silhouette supervision aligns the object shape.

\setlength\intextsep{-2pt}
\begin{wrapfigure}{r}{0.48\textwidth}
    \centering

    \resizebox{\linewidth}{!}{
    \begin{tabular}{@{}cccc@{}}
    \toprule
    \Cref{ssec:mcmc} & \textbf{Init CD $\downarrow$} & \textbf{Future CD $\downarrow$} & \textbf{MAE log E $\downarrow$} \\
    \midrule
    - & 10.79 & 223 & 1.10 \\
    \checkmark & \ \ 4.69 & \ \ 78 & 0.55 \\
    \bottomrule
    \end{tabular}
    }
    \captionof{table}{Effect of physics-aware geometric refinement on a representative scene.}
    \label{tab:geom_refinement}

    \vspace{0.5em}

    \includegraphics[width=\linewidth]{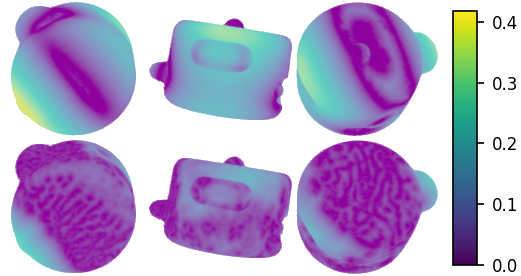}
    \captionof{figure}{Per-region geometric error before (\textit{top row}) and after (\textit{bottom row}) physics-aware refinement, projected onto the ground-truth mesh from multiple views.}
    \label{fig:error_map-n}
\end{wrapfigure}

\noindent\textbf{Geometry Refinement.}
We first illustrate how physics-aware geometry refinement improves the reconstruction quality of the initial geometry in the first frame.
To isolate geometric accuracy from scale ambiguity, we align both the initial and refined geometry to the true global scene scale before comparing them against the ground-truth point cloud using CD.
\cref{tab:geom_refinement} shows that refinement reduces not only the initial CD but also the future-prediction CD and the MAE of $\log E$.
\cref{fig:error_map-n} visualizes the per-region geometric error before (\textit{top row}) and after (\textit{bottom row}) refinement, showing reduced error across multiple views.
\setlength\intextsep{12pt plus 2pt minus 2pt}

\begin{figure}[t]
    \centering
    \subcaptionbox{Target frame}[0.25\linewidth]{%
        \includegraphics[width=\linewidth]{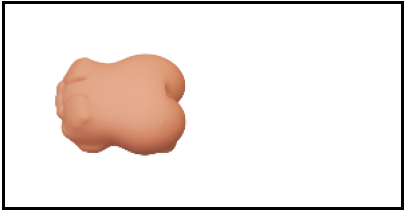}}
    \subcaptionbox{Rendered image}[0.25\linewidth]{%
        \includegraphics[width=\linewidth]{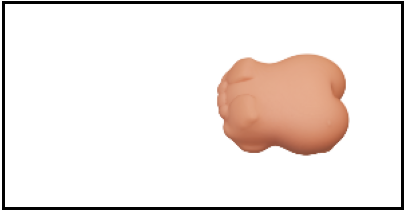}}
    \subcaptionbox{Position gradient map}[0.25\linewidth]{%
        \includegraphics[width=\linewidth]{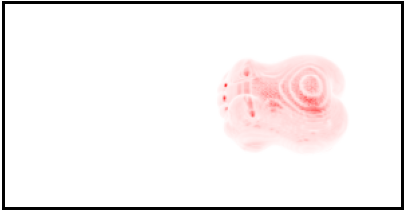}}
    \vspace{-1.5mm}
    \caption{
        Effect of the differentiable position map.
        The per-Gaussian position gradient $-\partial \mathcal{L}_{\text{sil}}/\partial \mathbf{x}_i$ points left (red), the correct direction toward the target shape (\cref{ssec:posmap}).
    }
    \label{fig:posmap_grad}
\end{figure}

\noindent\textbf{Differentiable Position Map.}
To visualize the gradients provided by the differentiable position map, \cref{fig:posmap_grad} renders the positional gradient $-\partial \mathcal{L}_{\text{sil}}/\partial \mathbf{x}_i$ of each Gaussian as an image, alongside the target and the current rendering.
The gradients point toward the target shape (leftward, shown in red).
Further details are provided in the supplementary material.

\section{Conclusion}
\label{sec:conclusion}
We present \textit{MonoPhysics}, a framework for monocular inverse physics of general deformable objects that jointly recovers geometry, appearance, and material parameters from a single-camera video.
Our key insight is that, without multi-view constraints, several paths between the visual and physical sides of the inverse physics pipeline become weak or missing.
Thus, we propose three visual-physical bridges: scene re-parameterization, physics-aware particle management, and a differentiable position map.
Together, these bridges yield more accurate future prediction than prior monocular methods.

Our method also has limitations that point to promising future directions.
First, our refinement reduces but does not eliminate geometric errors, and regions where neither visual nor physical signals are informative remain difficult.
Second, following prior work, our pipeline assumes known camera parameters, a known ground plane, and a known constitutive model, which limits its applicability to fully in-the-wild videos.
Third, all evaluated sequences are gravity-driven falls onto a flat plane, and other scenarios remain untested.
Because a ballistic trajectory under known gravity can make absolute scale observable, scale may be more difficult to recover in these other scenarios.
Relaxing these assumptions is a natural next step toward monocular inverse physics in the wild.

\acksection
This work is supported by a National Institute of Health (NIH) NIBIB project \#R21EB035832 and National Science Foundation (NSF) CAREER Award \#2543161.


%
%
\bibliographystyle{plain}
\bibliography{main}

@String(CVPR  = {IEEE Conf. Comput. Vis. Pattern Recog.})

@String(ICCV  = {Int. Conf. Comput. Vis.})

@String(ECCV  = {Eur. Conf. Comput. Vis.})

@String(NeurIPS = {Adv. Neural Inform. Process. Syst.})

@String(CVPR  = {CVPR})

@String(ICCV  = {ICCV})

@String(ECCV  = {ECCV})

@String(NeurIPS = {NeurIPS})

@article{liu2025physflow,
  title={Unleashing the Potential of Multi-modal Foundation Models and Video Diffusion for 4D Dynamic Physical Scene Simulation},
  author={Liu, Zhuoman and Ye, Weicai and Luximon, Yan and Wan, Pengfei and Zhang, Di},
  journal={CVPR},
  year={2025}
}

@InProceedings{gao2025fluidnexus,
    title     = {FluidNexus: 3D Fluid Reconstruction and Prediction from a Single Video},
    author    = {Gao, Yue and Yu, Hong-Xing and Zhu, Bo and Wu, Jiajun},
    booktitle = {Proceedings of the IEEE/CVF Conference on Computer Vision and Pattern Recognition (CVPR)},
    month     = {June},
    year      = {2025},
}

@INPROCEEDINGS{UniPhy,
  author={Mittal, Himangi and Zhuang, Peiye and Lee, Hsin-Ying and Tulsiani, Shubham},
  booktitle={2025 IEEE/CVF Conference on Computer Vision and Pattern Recognition (CVPR)}, 
  title={UniPhy: Learning a Unified Constitutive Model for Inverse Physics Simulation}, 
  year={2025},
  volume={},
  number={},
  pages={16208-16218},
  doi={10.1109/CVPR52734.2025.01511}
}

@InProceedings{SfC,
    author    = {Kaneko, Takuhiro},
    title     = {Structure from Collision},
    booktitle = {Proceedings of the IEEE/CVF Conference on Computer Vision and Pattern Recognition (CVPR)},
    month     = {June},
    year      = {2025},
    pages     = {16314-16324}
}

@InProceedings{vid2sim,
    author    = {Chen, Chuhao and Dou, Zhiyang and Wang, Chen and Huang, Yiming and Chen, Anjun and Feng, Qiao and Gu, Jiatao and Liu, Lingjie},
    title     = {Vid2Sim: Generalizable, Video-based Reconstruction of Appearance, Geometry and Physics for Mesh-free Simulation},
    booktitle = {Proceedings of the IEEE/CVF Conference on Computer Vision and Pattern Recognition (CVPR)},
    month     = {June},
    year      = {2025},
    pages     = {26545-26555}
}

@InProceedings{Lee_2024_CVPR,
    author    = {Lee, Joo Chan and Rho, Daniel and Sun, Xiangyu and Ko, Jong Hwan and Park, Eunbyung},
    title     = {Compact 3D Gaussian Representation for Radiance Field},
    booktitle = {Proceedings of the IEEE/CVF Conference on Computer Vision and Pattern Recognition (CVPR)},
    month     = {June},
    year      = {2024},
    pages     = {21719-21728}
}

@inproceedings{pun2025brickgpt,
    title     = {Generating Physically Stable and Buildable Brick Structures from Text},
    author    = {Pun, Ava and Deng, Kangle and Liu, Ruixuan and Ramanan, Deva and Liu, Changliu and Zhu, Jun-Yan},
    booktitle = {ICCV},
    year      = {2025}
}

@article{
    jiang2025phystwin,
    title={PhysTwin: Physics-Informed Reconstruction and Simulation of Deformable Objects from Videos},
    author={Jiang, Hanxiao and Hsu, Hao-Yu and Zhang, Kaifeng and Yu, Hsin-Ni and Wang, Shenlong and Li, Yunzhu},
    journal={ICCV},
    year={2025}
}

@InProceedings{MASIV,
    author    = {Zhao, Yizhou and Chen, Haoyu and Liu, Chunjiang and Li, Zhenyang and Herrmann, Charles and Hur, Junhwa and Li, Yinxiao and Yang, Ming-Hsuan and Raj, Bhiksha and Xu, Min},
    title     = {Toward Material-Agnostic System Identification from Videos},
    booktitle = {Proceedings of the IEEE/CVF International Conference on Computer Vision (ICCV)},
    month     = {October},
    year      = {2025},
    pages     = {5944-5956}
}

@inproceedings{PPR,
	title={Physically Plausible Reconstruction from Monocular Videos},
	author={Yang, Gengshan
	and Yang, Shuo
	and Zhang, John Z.
	and Manchester, Zachary
	and Ramanan, Deva},
	booktitle = {ICCV},
	year={2023},
}

@InProceedings{SpringGaus,
    author="Zhong, Licheng
    and Yu, Hong-Xing
    and Wu, Jiajun
    and Li, Yunzhu",
    editor="Leonardis, Ale{\v{s}}
    and Ricci, Elisa
    and Roth, Stefan
    and Russakovsky, Olga
    and Sattler, Torsten
    and Varol, G{\"u}l",
    title="Reconstruction and Simulation of Elastic Objects with Spring-Mass 3D Gaussians",
    booktitle="Computer Vision -- ECCV 2024",
    year="2025",
    publisher="Springer Nature Switzerland",
    address="Cham",
    pages="407--423",
    isbn="978-3-031-72627-9"
}

@InProceedings{SEA-RAFT,
    author="Wang, Yihan
    and Lipson, Lahav
    and Deng, Jia",
    editor="Leonardis, Ale{\v{s}}
    and Ricci, Elisa
    and Roth, Stefan
    and Russakovsky, Olga
    and Sattler, Torsten
    and Varol, G{\"u}l",
    title="SEA-RAFT: Simple, Efficient, Accurate RAFT for Optical Flow",
    booktitle="Computer Vision -- ECCV 2024",
    year="2025",
    publisher="Springer Nature Switzerland",
    address="Cham",
    pages="36--54",
    isbn="978-3-031-72667-5"
}

@InProceedings{Kaneko_2024_CVPR,
    author    = {Kaneko, Takuhiro},
    title     = {Improving Physics-Augmented Continuum Neural Radiance Field-Based Geometry-Agnostic System Identification with Lagrangian Particle Optimization},
    booktitle = {Proceedings of the IEEE/CVF Conference on Computer Vision and Pattern Recognition (CVPR)},
    month     = {June},
    year      = {2024},
    pages     = {5470-5480}
}

@inproceedings{
    gao2025seeing,
    title={Seeing the Wind from a Falling Leaf},
    author={Zhiyuan Gao and Jiageng Mao and Hong-Xing Yu and Haozhe Lou and Emily Yue-ting Jia and Jernej Barbic and Jiajun Wu and Yue Wang},
    booktitle={The Thirty-ninth Annual Conference on Neural Information Processing Systems},
    year={2025},
    url={https://openreview.net/forum?id=4NaW9mbTqq}
}

@inproceedings{GIC,
    author = {Cai, Junhao and Yang, Yuji and Yuan, Weihao and He, Yisheng and Dong, Zilong and Bo, Liefeng and Cheng, Hui and Chen, Qifeng},
    booktitle = {Advances in Neural Information Processing Systems},
    doi = {10.52202/079017-2388},
    editor = {A. Globerson and L. Mackey and D. Belgrave and A. Fan and U. Paquet and J. Tomczak and C. Zhang},
    pages = {75035--75063},
    publisher = {Curran Associates, Inc.},
    title = {GIC: Gaussian-Informed Continuum for Physical Property Identification and Simulation},
    url = {https://proceedings.neurips.cc/paper_files/paper/2024/file/89379d5fc6eb34ff98488202fb52b9d0-Paper-Conference.pdf},
    volume = {37},
    year = {2024}
}

@inproceedings{
    zhu2024motiongs,
    title={Motion{GS}: Exploring Explicit Motion Guidance for Deformable 3D Gaussian Splatting},
    author={Ruijie Zhu and Yanzhe Liang and Hanzhi Chang and Jiacheng Deng and Jiahao Lu and Wenfei Yang and Tianzhu Zhang and Yongdong Zhang},
    booktitle={The Thirty-eighth Annual Conference on Neural Information Processing Systems},
    year={2024},
    url={https://openreview.net/forum?id=6FTlHaxCpR}
}

@inproceedings{ni2024phyrecon,
  title={PhyRecon: Physically Plausible Neural Scene Reconstruction}, 
  author={Ni, Junfeng and Chen, Yixin and Jing, Bohan and Jiang, Nan and Wang, Bin and Dai, Bo and Li, Puhao and Zhu, Yixin and Zhu, Song-Chun and Huang, Siyuan},
  journal={Advances in Neural Information Processing Systems},
  year={2024}
}

@inproceedings{
    guo2024physically,
    title={Physically Compatible 3D Object Modeling from a Single Image},
    author={Minghao Guo and Bohan Wang and Pingchuan Ma and Tianyuan Zhang and Crystal Elaine Owens and Chuang Gan and Joshua B. Tenenbaum and Kaiming He and Wojciech Matusik},
    booktitle={The Thirty-eighth Annual Conference on Neural Information Processing Systems},
    year={2024},
    url={https://openreview.net/forum?id=k29Iv0XrBF}
}

@inproceedings{3DGS_MCMC,
 author = {Kheradmand, Shakiba and Rebain, Daniel and Sharma, Gopal and Sun, Weiwei and Tseng, Yang-Che and Isack, Hossam and Kar, Abhishek and Tagliasacchi, Andrea and Yi, Kwang Moo},
 booktitle = {Advances in Neural Information Processing Systems},
 doi = {10.52202/079017-2573},
 editor = {A. Globerson and L. Mackey and D. Belgrave and A. Fan and U. Paquet and J. Tomczak and C. Zhang},
 pages = {80965--80986},
 publisher = {Curran Associates, Inc.},
 title = {3D Gaussian Splatting as Markov Chain Monte Carlo},
 url = {https://proceedings.neurips.cc/paper_files/paper/2024/file/93be245fce00a9bb2333c17ceae4b732-Paper-Conference.pdf},
 volume = {37},
 year = {2024}
}

@inproceedings{
    qiao2022neuphysics,
    author  = {Qiao, Yi-Ling and Gao, Alexander and Lin, Ming C.},
    title  = {NeuPhysics: Editable Neural Geometry and Physics from Monocular Videos},
    booktitle = {Conference on Neural Information Processing Systems (NeurIPS)},
    year  = {2022},
}

@InProceedings{NCLaws,
  title = 	 {Learning Neural Constitutive Laws from Motion Observations for Generalizable {PDE} Dynamics},
  author =       {Ma, Pingchuan and Chen, Peter Yichen and Deng, Bolei and Tenenbaum, Joshua B. and Du, Tao and Gan, Chuang and Matusik, Wojciech},
  booktitle = 	 {Proceedings of the 40th International Conference on Machine Learning},
  pages = 	 {23279--23300},
  year = 	 {2023},
  editor = 	 {Krause, Andreas and Brunskill, Emma and Cho, Kyunghyun and Engelhardt, Barbara and Sabato, Sivan and Scarlett, Jonathan},
  volume = 	 {202},
  series = 	 {Proceedings of Machine Learning Research},
  month = 	 {23--29 Jul},
  publisher =    {PMLR},
  url = 	 {https://proceedings.mlr.press/v202/ma23a.html}
}

@inproceedings{
    xiong2025topogaussian,
    title={TopoGaussian: Inferring Internal Topology Structures from Visual Clues},
    author={Xiaoyu Xiong and Changyu Hu and Chunru Lin and Pingchuan Ma and Chuang Gan and Tao Du},
    booktitle={The Thirteenth International Conference on Learning Representations},
    year={2025},
    url={https://openreview.net/forum?id=B5PbOsJqt3}
}

@inproceedings{
    li2023pacnerf,
    title={{PAC}-Ne{RF}: Physics Augmented Continuum Neural Radiance Fields for Geometry-Agnostic System Identification},
    author={Xuan Li and Yi-Ling Qiao and Peter Yichen Chen and Krishna Murthy Jatavallabhula and Ming Lin and Chenfanfu Jiang and Chuang Gan},
    booktitle={The Eleventh International Conference on Learning Representations },
    year={2023},
    url={https://openreview.net/forum?id=tVkrbkz42vc}
}

@inproceedings{ym2025physic,
  author    = {Yalandur Muralidhar, Pradyumna and Xue, Yuxuan and Xie, Xianghui and Kostyrko, Margaret and Pons-Moll, Gerard},
  title     = {PhySIC: Physically Plausible 3D Human-Scene Interaction and Contact from a Single Image},
  journal   = {SIGGRAPH Asia 2025 Conference Papers},
  year      = {2025},
}

@article{simplicits,
    author = {Modi, Vismay and Sharp, Nicholas and Perel, Or and Sueda, Shinjiro and Levin, David I. W.},
    title = {Simplicits: Mesh-Free, Geometry-Agnostic Elastic Simulation},
    year = {2024},
    issue_date = {July 2024},
    publisher = {Association for Computing Machinery},
    address = {New York, NY, USA},
    volume = {43},
    number = {4},
    issn = {0730-0301},
    url = {https://doi.org/10.1145/3658184},
    doi = {10.1145/3658184},
    journal = {ACM Trans. Graph.},
    month = jul,
    articleno = {117},
    numpages = {11}
}

@Article{kerbl3Dgaussians,
      author       = {Kerbl, Bernhard and Kopanas, Georgios and Leimk{\"u}hler, Thomas and Drettakis, George},
      title        = {3D Gaussian Splatting for Real-Time Radiance Field Rendering},
      journal      = {ACM Transactions on Graphics},
      number       = {4},
      volume       = {42},
      month        = {July},
      year         = {2023},
      url          = {https://repo-sam.inria.fr/fungraph/3d-gaussian-splatting/}
}

@inproceedings{
    murthy2021gradsim,
    title={gradSim: Differentiable simulation for system identification and visuomotor control},
    author={J. Krishna Murthy and Miles Macklin and Florian Golemo and Vikram Voleti and Linda Petrini and Martin Weiss and Breandan Considine and J{\'e}r{\^o}me Parent-L{\'e}vesque and Kevin Xie and Kenny Erleben and Liam Paull and Florian Shkurti and Derek Nowrouzezahrai and Sanja Fidler},
    booktitle={International Conference on Learning Representations},
    year={2021},
    url={https://openreview.net/forum?id=c_E8kFWfhp0}
}

@inproceedings{
    Hu2020DiffTaichi:,
    title={DiffTaichi: Differentiable Programming for Physical Simulation},
    author={Yuanming Hu and Luke Anderson and Tzu-Mao Li and Qi Sun and Nathan Carr and Jonathan Ragan-Kelley and Fredo Durand},
    booktitle={International Conference on Learning Representations},
    year={2020},
    url={https://openreview.net/forum?id=B1eB5xSFvr}
}

@article{hong2025physics,
  title={Physics-Informed Deformable Gaussian Splatting: Towards Unified Constitutive Laws for Time-Evolving Material Field},
  author={Hong, Haoqin and Fan, Ding and Dou, Fubin and Zhou, Zhi-Li and Sun, Haoran and Zhu, Congcong and Chen, Jingrun},
  journal={arXiv preprint arXiv:2511.06299},
  year={2025}
}

@INPROCEEDINGS{GSO,
  author={Downs, Laura and Francis, Anthony and Koenig, Nate and Kinman, Brandon and Hickman, Ryan and Reymann, Krista and McHugh, Thomas B. and Vanhoucke, Vincent},
  booktitle={2022 International Conference on Robotics and Automation (ICRA)}, 
  title={Google Scanned Objects: A High-Quality Dataset of 3D Scanned Household Items}, 
  year={2022},
  volume={},
  number={},
  pages={2553-2560},
  doi={10.1109/ICRA46639.2022.9811809}
}

@InProceedings{sinkhorn,
  title = 	 {Interpolating between Optimal Transport and MMD using Sinkhorn Divergences},
  author =       {Feydy, Jean and S\'{e}journ\'{e}, Thibault and Vialard, Fran\c{c}ois-Xavier and Amari, Shun-ichi and Trouve, Alain and Peyr\'{e}, Gabriel},
  booktitle = 	 {Proceedings of the Twenty-Second International Conference on Artificial Intelligence and Statistics},
  pages = 	 {2681--2690},
  year = 	 {2019},
  editor = 	 {Chaudhuri, Kamalika and Sugiyama, Masashi},
  volume = 	 {89},
  series = 	 {Proceedings of Machine Learning Research},
  month = 	 {16--18 Apr},
  publisher =    {PMLR},
  url = 	 {https://proceedings.mlr.press/v89/feydy19a.html}
}

@article{NeRF,
    author = {Mildenhall, Ben and Srinivasan, Pratul P. and Tancik, Matthew and Barron, Jonathan T. and Ramamoorthi, Ravi and Ng, Ren},
    title = {NeRF: representing scenes as neural radiance fields for view synthesis},
    year = {2021},
    issue_date = {January 2022},
    publisher = {Association for Computing Machinery},
    address = {New York, NY, USA},
    volume = {65},
    number = {1},
    issn = {0001-0782},
    url = {https://doi.org/10.1145/3503250},
    doi = {10.1145/3503250},
    journal = {Commun. ACM},
    month = dec,
    pages = {99–106},
    numpages = {8}
}

@article{sam3dteam2025sam3d3dfyimages,
      title={SAM 3D: 3Dfy Anything in Images}, 
      author={SAM 3D Team and Xingyu Chen and Fu-Jen Chu and Pierre Gleize and Kevin J Liang and Alexander Sax and Hao Tang and Weiyao Wang and Michelle Guo and Thibaut Hardin and Xiang Li and Aohan Lin and Jiawei Liu and Ziqi Ma and Anushka Sagar and Bowen Song and Xiaodong Wang and Jianing Yang and Bowen Zhang and Piotr Dollár and Georgia Gkioxari and Matt Feiszli and Jitendra Malik},
      year={2025},
      eprint={2511.16624},
      archivePrefix={arXiv},
      primaryClass={cs.CV},
      url={https://arxiv.org/abs/2511.16624}, 
}

@article{rho2025projo4d,
  title   = {ProJo4D: Progressive Joint Optimization for Sparse-View Inverse Physics Estimation},
  author  = {Daniel Rho and Jun Myeong Choi and Biswadip Dey and Roni Sengupta},
  journal = {Transactions on Machine Learning Research},
  year    = {2026},
  month   = {05},
  url     = {https://openreview.net/forum?id=pqvVrqlXCZ}
}

\newpage

\appendix

\section{Implementation Details}
\label{sec:supple_impl}

\noindent\textbf{Optimization.}
We optimize the physical dynamics parameters for 250 iterations, using the same schedule for all constitutive models.
The number of simulated frames per iteration starts at 4 and increases linearly to the full sequence over the first 150 iterations.
The remaining iterations use the full sequence.
Within each iteration, we also optimize appearance-related parameters (excluding position and physical parameters) for 100 steps on randomly sampled frames.
After dynamics optimization, we refine appearance for an additional 5{,}000 iterations.
For real-world captures, all methods (including ours) are run for a total of 500 iterations, following the SpringGaus configuration.
For our method, we use the same curriculum as above, reaching the full sequence after 150 iterations.

\noindent\textbf{Loss Hyperparameters.}
As shown in \cref{eq:total_loss_1}, all loss terms are combined with unit weights.
For the particle distribution regularizer $\mathcal{L}_{\text{distr}}$, we use $K=3$ nearest neighbors per particle and set the lower and upper bounds to $\tau_{\min}=0.3$ and $\tau_{\max}=0.8$, respectively.
These bounds leave sufficient room for adaptive geometry while preventing overly aggressive regularization.

\noindent\textbf{Learning Rates.}
We use the Adam optimizer with fixed learning rates throughout optimization and apply the same hyperparameters across all synthetic scenes.
The global scale $s$ and the initial velocity $\mathbf{v}_0$ both use a learning rate of $0.02$.
For material parameters, we use $0.1$ for Young's modulus $E$, $0.01$ for Poisson's ratio $\nu$, and $0.05$ for yield stress $\sigma_y$.
For Gaussian features, we use $0.0002$ for position and (visual) Gaussian scale, and $0.01$ for color and opacity.
For real-world captures, we instead use $0.05$ for $\mathbf{v}_0$ and $\nu$, $0.01$ for $s$, and $0.0001$ for position and Gaussian scale, keeping all other learning rates unchanged.

\noindent\textbf{Computational Resources and Cost.}
All experiments are conducted on a single NVIDIA RTX A6000 GPU.
Optimization takes approximately 1--2 hours per run for both our method and the baselines.

\section{Physics-Aware Geometry Refinement}

\noindent\textbf{Per-particle physical volume.}
The detailed algorithm is described in \cref{alg:induced_vol}.
In our implementation, we run the iteration for 5 steps.
Given that 50 to 200 simulation steps are typically required between two consecutive frames in our datasets, this costs less than $1/10$ of a single forward simulation pass for one frame.

\noindent\textbf{Particle management.}
Every 10 iterations, we identify the bottom $1\%$ of Gaussians by either visual importance $p^{\text{vis}}_i$ or physical importance $p^{\text{phys}}_i$.
We replace them by deterministically selecting the same number of Gaussians with the highest overall importance $p_i$, splitting each in half, and relocating the resulting copies to the positions of the removed ones.
For relocation, we use the algorithm proposed in 3DGS-MCMC~\cite{3DGS_MCMC}.

\vspace{3em}
\begin{algorithm}[h]
    \caption{Iterative Volume Estimation}
    \label{alg:induced_vol}
    \begin{algorithmic}[1]
    \Require Positions $\{\mathbf{x}_i\}_{i=1}^N$, grid spacing $\Delta x$, iterations $T$
    \Statex \textbf{Notation:} $i \in \{1,\ldots,N\}$: particle index;\; $\ell$: grid node index
    \Statex \phantom{\textbf{Notation:}} $w_{i\ell} \geq 0$: B-spline weight from particle $i$ to node $\ell$;\; $\Delta x^3$: cell volume
    \Statex \textit{Initialization}
    \State $n_\ell \leftarrow \textstyle\sum_i w_{i\ell}$ \hfill \Comment{weighted particle count at node $\ell$}
    \State $\hat{n}_i \leftarrow \textstyle\sum_\ell w_{i\ell}\, n_\ell$ \hfill \Comment{weighted sum of node occupancies over all nodes influencing particle $i$}
    \State $V_i \leftarrow \min\left(\Delta x^3 / \hat{n}_i,\; \Delta x^3\right)$ \hfill \Comment{initial volume: one cell shared among $\hat{n}_i$ particles}
    \Statex \textit{Iterative refinement}
    \For{$t = 1, \ldots, T$}
        \State $\tilde{V}_\ell \leftarrow \textstyle\sum_i w_{i\ell}\, V_i$ \hfill \Comment{total volume attributed to node $\ell$}
        \State $c_\ell \leftarrow \min\left(\Delta x^3 / \tilde{V}_\ell,\; 1\right)$ \hfill \Comment{correction: scale down overfull nodes, leave underfull ones}
        \State $V_i \leftarrow \min\left(V_i \cdot \textstyle\sum_\ell w_{i\ell}\, c_\ell,\; \Delta x^3\right)$ \hfill \Comment{update $V_i$ by gathering node corrections}
    \EndFor
    \State \Return $\{V_i\}_{i=1}^N$
    \end{algorithmic}
\end{algorithm}

\section{Differentiable Position Map}
\label{sec:supple_posmap}
We expand on the position-map reparameterization introduced in \cref{ssec:posmap}, showing why standard 3DGS rendering does not provide pixel-coordinate gradients and how the proposed reparameterization enables them.

\noindent\textbf{Standard 3DGS rendering provides no positional gradient from pixel-coordinate losses.}
In \cref{eq:3dgs_render}, the rendered color at pixel $\mathbf{p}$ is a function of the per-Gaussian quantities $\{\mathbf{c}_i, o_i, \mathbf{x}_i^{2D}, \boldsymbol{\Sigma}_i^{2D}\}$, while $\mathbf{p}$ itself is a fixed integer pixel index.
Therefore, for any loss $\mathcal{L}(\mathbf{p})$ defined on pixel \emph{coordinates} rather than rendered colors,
\begin{equation}
\label{eq:supple_zerograd}
    \frac{\partial \mathcal{L}(\mathbf{p})}{\partial \mathbf{x}_i^{2D}} = 0.
\end{equation}
In other words, standard rasterization provides no gradients from a pixel-coordinate loss to any Gaussian's image-space position.

\noindent\textbf{The differentiable position map provides positional gradients from pixel-coordinate losses.}
For a pixel $\mathbf{p}$ covered by Gaussian $i$, the differentiable position from \cref{eq:reparam} is
\begin{equation*}
    \tilde{\mathbf{p}}_i = \mathbf{x}_i^{2D} + (\boldsymbol{\Sigma}_i^{2D})^{1/2} \mathbf{z}, \qquad \mathbf{z} = \mathrm{sg}\!\left((\boldsymbol{\Sigma}_i^{2D})^{-1/2}(\mathbf{p} - \mathbf{x}_i^{2D})\right).
\end{equation*}
In the forward pass, $(\boldsymbol{\Sigma}_i^{2D})^{1/2} \mathbf{z} = \mathbf{p} - \mathbf{x}_i^{2D}$, so $\tilde{\mathbf{p}}_i = \mathbf{p}$ numerically.
That is, the rendered position map equals the pixel grid.
However, because $\mathbf{z}$ is treated as a constant under the stop-gradient,
\begin{equation}
\label{eq:supple_unitgrad}
    \frac{\partial \tilde{\mathbf{p}}_i}{\partial \mathbf{x}_i^{2D}} = \mathbf{I}.
\end{equation}
A pixel-coordinate loss applied to $\tilde{\mathbf{p}}_i$ therefore passes its gradient directly to the 2D mean of the covering Gaussian, from which it propagates through the projection and the differentiable simulation back to the Gaussians' 3D positions and physical state.

\section{Dataset Details}
\label{sec:supple_dataset}

\noindent\textbf{Scene composition.}
Our synthetic dataset is built from 10 source meshes drawn from Google Scanned Objects (GSO)~\cite{GSO}: 5 objects simulated with an elastic (Neo-Hookean) constitutive model and 5 simulated with a plasticine constitutive model.
For each scene, the material parameters (Young's modulus $E$, Poisson's ratio $\nu$, and yield stress $\sigma_y$ for plasticine) and the initial velocity are independently sampled at random.
For elastic objects, we sample $\log_{10} E$ from $(3.5, 7.0)$ and $\nu$ from $(0, 0.49)$.
For plasticine objects, we sample $\log_{10} E$ from $(4.0, 5.5)$, $\nu$ from $(0.2, 0.45)$, and $\log_{10} \sigma_y$ from $(0.1, 1.5)$.
All scenes share a fixed density of $1000$\,kg/m$^3$, gravity  of 9.81 m/s$^2$, and a frictionless ground plane at $z=0$.
Each sequence is simulated with MPM at 20 fps.

\noindent\textbf{Scenes.}
\cref{tab:supple_dataset_scenes} lists the 10 source GSO objects and the constitutive model used for each.

\begin{table}[h]
\centering
\caption{Source GSO objects and their assigned constitutive model.}
\label{tab:supple_dataset_scenes}
\begin{tabular}{ll}
\toprule
\textbf{Object} & \textbf{Constitutive model} \\
\midrule
BIRD\_RATTLE            & Neo-Hookean \\
CHICKEN\_NESTING        & Neo-Hookean \\
PEEKABOO\_ROLLER        & Neo-Hookean \\
TWISTED\_PUZZLE         & Neo-Hookean \\
WHALE\_WHISTLE\_6PCS\_SET & Neo-Hookean \\
\midrule
BABY\_CAR         & Plasticine \\
COAST\_GUARD\_BOAT & Plasticine \\
LACING\_SHEEP     & Plasticine \\
MINI\_EXCAVATOR   & Plasticine \\
OWL\_SORTER       & Plasticine \\
\bottomrule
\end{tabular}
\end{table}
\vspace{2em}

\noindent\textbf{Constitutive models.}
Both material models use the same conversion from $E$ and $\nu$ to the Lam\'e parameters:
\begin{equation}
\mu = \frac{E}{2(1+\nu)}, \qquad \lambda = \frac{E\,\nu}{(1+\nu)(1-2\nu)}.
\end{equation}

\noindent\emph{Neo-Hookean (elastic).}
We use the compressible Neo-Hookean Kirchhoff stress:
\begin{equation}
\boldsymbol{\tau} = \mu\,\mathbf{F}\mathbf{F}^{\!\top} + \bigl(\lambda \ln J - \mu\bigr)\mathbf{I}, \qquad J = \det \mathbf{F},
\end{equation}
where $\mathbf{F}$ is the deformation gradient.

\noindent\emph{Plasticine.}
The plasticine model combines a Fixed Corotated elastic stress with a von Mises return mapping in principal log-strain space.
Using the polar decomposition $\mathbf{F} = \mathbf{R}\mathbf{S}$, the elastic Kirchhoff stress is
\begin{equation}
\boldsymbol{\tau} = 2\mu\,(\mathbf{F} - \mathbf{R})\mathbf{F}^{\!\top} + \lambda\,J\,(J - 1)\,\mathbf{I}.
\end{equation}
Plasticity is enforced by a return mapping on the trial elastic deformation gradient.
Given the SVD $\mathbf{F}^{\text{trial}} = \mathbf{U}\boldsymbol{\Sigma}\mathbf{V}^{\!\top}$, we compute the principal log strains $\boldsymbol{\varepsilon} = \ln \boldsymbol{\Sigma}$ and their deviatoric part $\hat{\boldsymbol{\varepsilon}} = \boldsymbol{\varepsilon} - \tfrac{1}{3}\,\mathrm{tr}(\boldsymbol{\varepsilon})\,\mathbf{I}$.
The von Mises yield indicator is
\begin{equation}
y = \|\hat{\boldsymbol{\varepsilon}}\| - \frac{\sigma_y}{2\mu}.
\end{equation}
If $y > 0$, we project $\boldsymbol{\varepsilon} \leftarrow \boldsymbol{\varepsilon} - y\,\hat{\boldsymbol{\varepsilon}} / \|\hat{\boldsymbol{\varepsilon}}\|$ and update $\mathbf{F} \leftarrow \mathbf{U}\exp(\boldsymbol{\varepsilon})\mathbf{V}^{\!\top}$.
Otherwise, $\mathbf{F}$ is left unchanged.

\section{Full Results on Our Synthetic Dataset}
\label{sec:supple_full_results}

Due to space, \cref{tab:main_table_Ours} in the main paper reports medians only.
\cref{tab:supple_main_table_Ours_iqr} provides the full results as median [Q1, Q3].

\begin{table*}[h]
    {\captionsetup{position=top}
    \captionof{table}{
        \textbf{Full evaluation on our synthetic dataset.}
        We evaluate 3D trajectory accuracy and rendering quality on future frames.
        We also evaluate physical parameter estimation.
        Values are median [Q1, Q3].
        This extends \cref{tab:main_table_Ours}, which reports medians only due to space.
        Bold marks the best method in each column, excluding \textit{Oracle Midpoint}.
    }
    \label{tab:supple_main_table_Ours_iqr}
    \centering
    \resizebox{\linewidth}{!}{%
        \begin{tabular}{lccccccccc}
        \toprule
        & \multicolumn{2}{c}{\textbf{3D Trajectory}} & \multicolumn{3}{c}{\textbf{Future Frame Rendering}} &\multicolumn{4}{c}{\textbf{Physical Parameter Estimation}} \\
         
        \cmidrule(lr){2-3} \cmidrule(lr){4-6} \cmidrule(lr){7-10}
        & \textbf{CD $\downarrow$} & \textbf{EMD $\downarrow$} 
        & \textbf{PSNR $\uparrow$}  & \textbf{LPIPS $\downarrow$} & \textbf{SSIM $\uparrow$}  
         & \textbf{MAE $\mathbf{v}_0$ $\downarrow$} 
        & \textbf{MAE $\log E$ $\downarrow$} & \textbf{MAE $\nu$ $\downarrow$} 
        & \textbf{MAE $\log \sigma_y$ $\downarrow$} \\
        \midrule

        \multicolumn{9}{l}{\textbf{Elastic}} \\
        \midrule
        \multicolumn{2}{l}{\textit{Oracle Midpoint}} & & & & & 0.15 [0.14, 0.31] & 1.11 [0.93, 1.23] & 0.20 [0.15, 0.21] & - \\
        PAC-NeRF & 438 [199, \ \ 696] & 0.74 [0.54, 0.88] & 17.4 [16.5, 19.0] & 0.19 [0.16, 0.22] & 0.92 [0.91, 0.93] & \textbf{0.04} [0.03, 0.10] & 1.33 [0.43, 2.27] & \textbf{0.06} [0.03, 0.26] & - \\
        SpringGaus & 693 [592, \ \ 814] & 0.97 [0.89, 1.05] & 16.7 [15.1, 18.4] & 0.24 [0.20, 0.28] & 0.91 [0.90, 0.92] & 0.16 [0.10, 0.25] & - & - & - \\
        GIC & 421 [174, 1122] & 0.75 [0.47, 1.16] & 18.5 [17.7, 19.9] & 0.18 [0.16, 0.23] & 0.92 [0.91, 0.93] & 0.09 [0.05, 0.17] & 0.66 [0.30, 1.17] & 0.21 [0.17, 0.29] & - \\
        \textbf{Ours} & \ \ \textbf{24} [\ \ 11, \ \ \ \ 77] & \textbf{0.26} [0.16, 0.38] & \textbf{21.1} [19.3, 22.1] & \textbf{0.12} [0.10, 0.15] & \textbf{0.93} [0.92, 0.94] & 0.09 [0.05, 0.14] & \textbf{0.31} [0.19, 0.49] & 0.07 [0.02, 0.09] & - \\

        \midrule
        
        \multicolumn{9}{l}{\textbf{Plasticine}} \\
        \midrule
        \multicolumn{2}{l}{\textit{Oracle Midpoint}} & & & & & 0.22 [0.13, 0.23] & 0.32 [0.20, 0.65] & 0.01 [0.01, 0.04] & 0.81 [0.37, 0.86] \\
        PAC-NeRF & 247 [135, \ \ 433] & 0.62 [0.47, 0.75] & 19.8 [16.8, 22.9] & 0.14 [0.12, 0.20] & 0.94 [0.91, 0.96] & 0.14 [0.08, 0.20] & 2.55 [1.10, 3.39] & 0.17 [0.12, 0.24] & 0.57 [0.49, 1.06] \\
        SpringGaus & 450 [223, \ \ 804] & 0.67 [0.55, 0.93] & 17.3 [16.4, 20.2] & 0.20 [0.16, 0.26] & 0.92 [0.90, 0.94] & 0.13 [0.07, 0.22]  & - & - & - \\
        GIC & 333 [185, \ \ 509] & 0.65 [0.53, 0.75] & 22.0 [19.9, 24.4] & 0.14 [0.12, 0.20] & 0.94 [0.92, 0.96] & \textbf{0.12} [0.07, 0.27] & 0.95 [0.45, 2.06] & 0.07 [0.04, 0.27] & 0.34 [0.25, 0.65] \\
        \textbf{Ours} & \ \ \textbf{19} [\ \ \ \ 7, \ \ \ \ 72] & \textbf{0.24} [0.13, 0.43] & \textbf{23.4} [22.4, 25.4] & \textbf{0.09} [0.07, 0.11] & \textbf{0.95} [0.93, 0.96] & 0.14 [0.10, 0.23] & \textbf{0.40} [0.14, 0.81] & \textbf{0.05} [0.02, 0.14] & \textbf{0.33} [0.19, 0.50] \\
        \bottomrule
        \end{tabular}%
    }
}
\end{table*}

\section{Details on Losses}
\label{sec:supple_details}

\noindent\textbf{Optical Flow Loss $\mathcal{L}_{flow}$}
The probability-based optical flow loss in the main paper uses a per-pixel uncertainty $\sigma$ produced by the flow estimator~\cite{SEA-RAFT}.
The full loss with explicit uncertainty is:
\begin{equation}
\label{eq:flow_full}
\begin{split}
    \mathcal{L}_{\text{flow}} = -\frac{1}{|\Omega_t|}\sum_{j \in \Omega_t}\!\Big[
        &\log p\!\left(\mathbf{f}_{0 \to t}(\mathbf{p}_j) \mid \hat{\mathbf{f}}_{0 \to t}(\mathbf{p}_j),\, \sigma_{0 \to t}(\mathbf{p}_j)\right) \\
        + &\log p\!\left(\mathbf{f}_{t\text{-}1 \to t}(\mathbf{p}_j) \mid \hat{\mathbf{f}}_{t\text{-}1 \to t}(\mathbf{p}_j),\, \sigma_{t\text{-}1 \to t}(\mathbf{p}_j)\right)\Big],
\end{split}
\end{equation}
where $\sigma_{i \to j}(\cdot)$ is the per-pixel uncertainty from frame $i$ to frame $j$ at pixel $\mathbf{p}$.
The likelihood downweights high-uncertainty pixels, which improves robustness when physical dynamics change visibility between frames.

\section{Broader Impacts}
\label{sec:supple_impact}
Recovering physical properties from monocular video can benefit robotic manipulation of deformable objects and the construction of digital twins and virtual environments.
However, the same capability could also be misused, for example, to create physically plausible but fake videos, or digital twins of objects and scenes captured without consent.
At present, our method's assumptions limit its direct application to in-the-wild captures, which reduces the immediate risk of such misuse.
Nevertheless, as monocular inverse physics matures, it should be used responsibly and with caution regarding these applications.


\end{document}